\documentclass[10pt,twocolumn,letterpaper]{article}

\usepackage{cvpr}
\usepackage{times}
\usepackage{epsfig}
\usepackage{graphicx}
\usepackage{amsmath}
\usepackage{amssymb}
\usepackage{subfigure}
\usepackage{xcolor}
\usepackage{footnote}

\usepackage{algorithm}
\usepackage{algpseudocode}

\graphicspath{{./figures/}{./figures/draw/}}

\def\x{{\mathbf x}}

\def\w{{\mathbf w}}
\def\h{{\mathbf h}}
\def\cc{{\mathbf c}}
\def\bb{{\mathbf b}}
\def\I{{\mathbf 1}}
\def\q{{\mathbf q}}

\def\B{{\mathbf B}}
\def\X{{\mathbf X}}
\def\Y{{\mathbf Y}}
\def\R{{\mathbb R}}
\def\H{{\mathbf H}}
\def\Z{{\mathbf Z}}
\def\W{{\mathbf W}}
\def\V{{\mathbf V}}
\def\Q{{\mathbf Q}}
\def\I{{\mathbf I}}
\def\1{{\mathbf 1}}
\def\S{{\mathbf S}}
\def\V{{\mathbf V}}

\newcommand\norm[1]{\left\lVert#1\right\rVert}
\usepackage[pagebackref=true,breaklinks=true,letterpaper=true,colorlinks,bookmarks=false]{hyperref}

\cvprfinalcopy % *** Uncomment this line for the final submission

\def\cvprPaperID{****} % *** Enter the CVPR Paper ID here

\ifcvprfinal\fi
\begin{document}

%%%%%%%%% TITLE
\title{Discrete Hashing with Deep Neural Network}
%\title{Learning Hash Codes with Binary Autoencoder}
\author{Thanh-Toan Do
\and
Anh-Zung Doan
%{\tt\small quang\_tran@sutd.edu.sg}
\and
Ngai-Man Cheung\\
%{\tt\small ngaiman\_cheung@sutd.edu.sg}\\
\and Singapore University of Technology and Design \\
% 8 Somapah Road, Singapore 487372\\
{\tt\small \{thanhtoan\_do, dung\_doan, ngaiman\_cheung\}@sutd.edu.sg}\\
}

\maketitle
%\thispagestyle{empty}

%%%%%%%%% ABSTRACT
\begin{abstract}
%This paper addresses the problem of learning binary hash codes for large scale image search. 
%We propose a novel hashing method based on deep neural network which allows to seek multiple hierarchical non-linear transformations to learn binary codes. The advantage of our deep model over previous deep model used in hashing is that our model contains necessary criteria for producing good codes such as similarity preserving, balance and independence.
% 
%One of difficulty problems in binary hashing is handling with binary constraints. Instead of relaxing this constraint during the learning binary codes as most previous works, in this paper, by introducing the auxiliary variable, we reformulate the optimization into two sub-optimization steps allowing us to directly solve binary constraints with out any relaxation. %One optimizes hash function, and one optimizes binary codes. This allows us to directly solve binary constraints with out any relaxation. 
%
%Extensive experiments on standard datasets show the proposed methods outperform or are competitive with state-of-the-art hashing methods. 

This paper addresses the problem of learning binary hash codes for large scale image search by proposing a novel hashing method based on deep neural network. The advantage of our deep model over previous deep model used in hashing is that our model contains necessary criteria for producing good codes such as similarity preserving, balance and independence. Another advantage of our method is that instead of relaxing the binary constraint of codes during the learning process as most previous works, in this paper, by introducing the auxiliary variable, we reformulate the optimization into two sub-optimization steps allowing us to efficiently solve binary constraints without any relaxation. 

The proposed method is also extended to the supervised hashing by leveraging the label information such that the learned binary codes preserve the pairwise label of inputs.

The experimental results on three benchmark datasets show the proposed methods outperform state-of-the-art hashing methods. 
\end{abstract}

%%%%%%%%% BODY TEXT
%-------------------------------------------------------------------------
\section{Introduction}
%Large scale visual search has attracted considerable attention because of easy availability of huge amounts of data. 
Large scale visual search has attracted attention because of easy availability of huge amounts of data also its wide applications~\cite{DBLP:journals/csur/DattaJLW08}. 
Two main difficulties when dealing with large scale visual search are efficient storage and fast searching. 
An attractive approach for handling those difficulties is binary hashing where each original high dimensional vector $\x \in \R^D$ is mapped to a binary low dimensional vector $\bb \in \R^L$ where $L\ll D$. The resulted binary vectors will allow the efficient storage. Furthermore, while the searching in original space costs $\mathcal{O}(ND)$ where $N$ is database size, the searching in binary space costs $\mathcal{O}(NL)$ with much smaller constant factor. This is because the hardware can efficiently compute the distance between data points in binary space (e.g. using XOR operator) and the entire dataset ($NL$ bits) can fit in the main memory. There is a wide range of hashing methods proposed in the literature~\cite{Grauman_review,DBLP:journals/corr/WangSSJ14}. They can be divided into two categories, i.e., data-independent and data-dependent.

Most methods in data-independent category rely on random projections for generating hash functions. The representatives for this category are Locality-Sensitive Hashing (LSH)~\cite{lsh_vldb09} and its extensions which extend Euclidean distance to other distances such as kernelized LHS~\cite{KLSH_iccv09,KLSH_nips09}, LSH with Mahalanobis distance~\cite{DBLP:journals/pami/KulisJG09}. 

Instead of using random projections, data-dependent category uses available training data for learning hash functions in unsupervised or supervised way. The representatives for this category include unsupervised hashing such as Spectral Hashing~\cite{DBLP:conf/nips/WeissTF08}, Iterative Quantization (ITQ)~\cite{DBLP:conf/cvpr/GongL11}, K-means Hashing~\cite{DBLP:conf/cvpr/HeWS13}, Spherical Hashing~\cite{CVPR12:SphericalHashing}, Isotropic Hashing~\cite{conf/nips/KongL12}~\textit{etc.}, and supervised hashing such as LDA Hashing~\cite{DBLP:journals/pami/StrechaBBF12}, %Semi-Supervised Hashing(SSH)~\cite{DBLP:journals/pami/WangKC12}, 
 Minimal Loss Hashing~\cite{DBLP:conf/icml/NorouziF11,DBLP:conf/nips/0002FS12}, ITQ-CCA~\cite{DBLP:conf/cvpr/GongL11}, FastHash~\cite{CVPR2014Lin}, Binary Reconstructive Embedding~\cite{Kulis_learningto},~\textit{etc.}

One of difficult problems in hashing is to design hash function which can capture nonlinear structures in input space. Most aforementioned methods assumed hashing functions as linear functions so they may not well capture the nonlinear manifold structure of inputs. Although several kernel-based hashing methods have proposed~\cite{CVPR12:Hashing,KLSH_iccv09,KLSH_nips09,DBLP:journals/pami/GongLGP13}, they suffer from scalability problem. 

Another difficult problem in hashing is to deal with binary constraint on codes. 
In general, the binary constraint imposed on the output of hash functions leads to mixed-integer optimization problem which is NP-hard. To handle with this difficulty, most aforementioned methods relax the constraint during learning process. With this relaxation, the continuous codes are learned first, they then are binarized (e.g. by thresholding or with an optimal rotation). This relaxation greatly simplifies the original binary constraint problem and its solution is suboptimal, i.e., the binary codes resulting of thresholded continuous codes is not necessary same as binary codes resulting by directly solving the thresholding in the learning process. 

\subsection{Related work} 
In order to better capture nonlinear manifold structure of inputs, there are few of hashing methods~\cite{DBLP:SeH,Liong_2015_CVPR,BA_CVPR15} relying on deep learning techniques. Semantic hashing~\cite{DBLP:SeH} is the first work using deep learning for hashing. Their model is formed by stacked of Restricted Boltzmann Machine and a pretraining step is required to train the model. In~\cite{BA_CVPR15}, the authors use linear autoencoder as hash functions seeking to reconstruct an input from the binary code produced by hidden layer of the network. Because the model in~\cite{BA_CVPR15} only uses shallow network (i.e. only one hidden layer) with linear activation function, it may not well capture nonlinear structure of inputs. In~\cite{Liong_2015_CVPR}, the authors use a deep neural network as hash functions. However, their unsupervised hashing method does not have the similarity preserving property that is not only similar inputs should likely have similar binary codes but also different inputs should likely have different binary codes. The similarity preserving property has been indicated as an important criterion for the hashing methods~\cite{DBLP:conf/nips/WeissTF08}.  

In order to handle with the binary constraint, semantic hashing~\cite{DBLP:SeH} and deep hashing~\cite{Liong_2015_CVPR} first solve in learning process the relaxed problem by discarding the constraint and then threshold the solved continuous solution, resulting the binary solution. Opposite to~\cite{DBLP:SeH,Liong_2015_CVPR}, linear binary autoencoder-based hashing~\cite{BA_CVPR15} directly solves binary constraint during learning process. They used an exhausted search (i.e., searching in $2^L$ solutions) for finding the best binary code which minimizes the objective function (the reconstruction error). This may cause the training process time-consuming when large number of bits is used to encode a sample. 
Recently, in supervised discrete hashing (SDH)~\cite{Shen_2015_CVPR}, the authors proposed a new method named~\textit{discrete cyclic coordinate descent} which efficiently solves the binary constraint without the relaxation. By solving the binary constraint bit by bit, they achieved an analytic solution for the processed bit. This makes the training process very efficient. It is worth noting that the objective function of SDH~\cite{Shen_2015_CVPR} is designed by basing on the assumption that the good hash codes are optimal for linear classification. This assumption may not be directly involved to the retrieval problem. %We also give an analysis on this problem in the section~\ref{subsec:sdh11-sdh2}. 

\subsection{Contribution}
In this work, we first propose a novel unsupervised hashing method based on deep learning techniques. By using deep neural network with nonlinear activation functions, our method can capture complex structure in inputs. Our objective function includes the criteria~\cite{DBLP:conf/nips/WeissTF08} for producing good binary codes such as similarity preserving, independent and balancing properties. This is different from~\cite{Liong_2015_CVPR} where only independent and balancing properties are considered. Furthermore, instead of doing relaxation when dealing with the binary constraint as previous works~\cite{Liong_2015_CVPR}, we directly solve the binary constraint during learning process, resulting binary codes of better quality. The main differences between our hashing method and recent deep learning-based unsupervised hashing Deep Hash (DH)~\cite{Liong_2015_CVPR} and linear Binary Autoencoder (BA)~\cite{BA_CVPR15} are summarized in Table~\ref{tab:benchmark}. The compared criteria are: is network-model deep?  Does the objective function consider the similarity preserving/independent/balancing of binary codes? How are the binary constraint on codes solved in the learning process? 

\begin{savenotes}
\begin{table}[!t]
\small
\centering
\caption{The difference between our method and deep learning-based unsupervised hashing~\cite{Liong_2015_CVPR,BA_CVPR15}.}
\label{tab:benchmark}
\begin{tabular}{|c|c|c|c|}
\hline
 		    &DH~\cite{Liong_2015_CVPR}  &BA~\cite{BA_CVPR15}  &Ours  \\ \hline	
			
Is model deep?	    &Yes					&No					  &Yes	  \\ \hline
Similarity preserving? &No					&Yes				  &Yes    \\ \hline
Independence? 	    &Yes\footnote{Although authors of Deep Hashing~\cite{Liong_2015_CVPR} considered the independent property in their objective function, they did the relaxation by putting the independent property on the weights of the network. It is different from us where the independent property is directly considered on the codes.}					&No					  &Yes    \\ \hline
Balance?			&Yes					&No					  &Yes	  \\ \hline
How to solve        &Relaxation             &Exhausted   		  &Closed- \\
binary const.?       &						&search				  &form    \\\hline
\end{tabular}
\end{table}
\end{savenotes}
After introducing the new method for unsupervised hashing, we then extend our method to supervised hashing by leveraging the label information such that the binary codes preserve the semantic (label) similarity between samples. Our main contributions are summarized as follows.
\begin{itemize}
\item We proposed a novel deep learning-based hashing method which allows to produce binary codes having expected properties such as similarity preserving, independent and balancing. 
\item We directly solve binary constraint during the learning process. The idea is to adaptly use the regularization approach~\cite{malick:hal-00389552} and recent proposed method~\textit{discrete cyclic coordinate descent}~\cite{Shen_2015_CVPR}.
\item The proposed method is first evaluated in unsupervised hashing setting. After that, we extend it to supervised hashing setting by leveraging the label information. 
%\item The extensive experiments on three benchmark datasets are provided. Experimental results shows the improvement of proposed method over several state-of-the-art hashing methods. 
\item The extensive results on three benchmark datasets show the improvement of proposed method over several state-of-the-art hashing methods. 
\end{itemize}
The remaining of this paper is organized as follows. Section~\ref{sec:udh-dnn} presents our proposed method for unsupervised hashing. Section~\ref{sec:eva_udh-dnn} evaluates the proposed unsupervised hashing method. Section~\ref{sec:sdh-dnn} presents our proposed method for supervised hashing. Section~\ref{sec:eva_sdh-dnn} evaluates the proposed supervised hashing. Section~\ref{sec:conclusion} concludes the paper. 
\section{Unsupervised Discrete Hashing with Deep Neural Network (UDH-DNN)}
\label{sec:udh-dnn}
\subsection{Formulation of UDH-DNN}
\label{subsec:formular_un}
Let $\X = \{\x_i\}_{i=1}^{m} \in \R^{D\times m}$ be set of $m$ training samples; each column of $\X$ corresponds to one sample. We target to learn the binary codes for each sample. Let $\B = \{\bb_i\}_{i=1}^{m} \in \R^{L\times m}$ be binary code matrix of $\X$; $L$ is the number of desire bits to encode a sample. In our work, the hash functions are defined as a deep neural network having $n$ layers (including input and output layers). %The output of layer $n-1$ can be used as binary codes. 

Let $s_l$ be number of units in layer $l$; $f^{(l)}$ be activation function of layer $l$; $\H^{(l)} = [\h_1^{(l)},\cdots,\h_m^{(l)}] \in \R^{s_l \times m}$ be output values of layer $l$ (for clarifying in later sections, we use $\H^{(1)} = \X$); $\W^{(l)} \in \R^{s_{l+1}\times s_l}$ be weight matrix connecting layer $l+1$ and layer $l$; $\cc^{(l)} \in \R^{s_{l+1}}$ be bias vector for units in layer $l+1$.

Our idea is to learn a deep neural network such that the sign of output values of layer $n-1$ can be used as binary codes and those codes %should have the similarity preserving property.
should give a good reconstruction of input. To achieve this goal, we choose to optimize the following objective function
\small
\begin{eqnarray}
\min_{\W,\cc} J &=& \frac{1}{2m} \norm{\X-\W^{(n-1)}sgn(\H^{(n-1)})-\cc^{(n-1)}\1_{1\times m}}^2 \nonumber \\ 
{}&&+\frac{\lambda_1}{2}\sum_{l=1}^n \norm{\W^{(l)}}^2 \label{eq:obj_ori}
\end{eqnarray}
\normalsize
where $\1_{1\times m}$ is a row vector having all elements equals to 1. In our formulation~(\ref{eq:obj_ori}), the binary code $\B$ is defined as $\B=sgn(\H^{(n-1)})$. 

The first term of the objective function~(\ref{eq:obj_ori}) makes sure that the binary code $\B$  gives a good reconstruction error of $\X$. It is worth noting that the reconstruction criterion does not directly measure the similarity preservation, %but instead preservers manifold structure of the inputs in some sense. The reason 
but it has been indicated in deep learning-based hashing methods~\cite{BA_CVPR15,DBLP:SeH} 
%\cite{BA_CVPR15,DBLP:SeH,DBLP:conf/esann/KrizhevskyH11} 
that the hash function defined by the neural networks containing reconstruction criterion can capture the data manifolds in a smooth way and indirectly preserve the similarity, encouraging (dis)similar inputs have to (dis)similar codes.
%Because the objective function minimizes the reconstruction error, given learned $\{\W^{(l)},\cc^{(l)}\}_{l=1}^{n-1}$ (by an optimizer), the network will output similar codes for similar inputs and different codes for different inputs. This implicates the similarity preserving property of codes. 
The second term is a regularization term that tends to decreases the magnitude of the weights, and helps to prevent the overfitting\footnote{As noted by Ng~\cite{Ng_deep}, the regularization is not usually applied to the bias terms $\cc$. Applying the regularization to the bias usually makes only a small difference to the final network.}. It is worth noting in~(\ref{eq:obj_ori}) that if we replace $sgn(\H^{(n-1)})$ by $\H^{(n-1)}$, the objective function~(\ref{eq:obj_ori}) can be seen as a deep autoencoder with linear decoder layer (i.e. the last layer $n$ uses linear activation function). 

Equivalently, by introducing the auxiliary variable $\B$, the objective function~(\ref{eq:obj_ori}) can be rewritten as 
\begin{eqnarray}
\min_{\W,\cc,\B} J &=& \frac{1}{2m} \norm{\X-\W^{(n-1)}\B-\cc^{(n-1)}\1_{1\times m}}^2 \nonumber \\ 
{}&&+\frac{\lambda_1}{2}\sum_{l=1}^n \norm{\W^{(l)}}^2 \label{eq:obj_2}
\end{eqnarray}
\hspace{0.3cm} s.t. \\
\vspace{-0.5cm}
\begin{equation}
\B = sgn(\H^{(n-1)}) \label{eq:binary}
\end{equation} 
The benefit of introducing the auxiliary variable $\B$ is that we can decompose the difficult optimization problem~(\ref{eq:obj_ori}) into two sub optimization problems where we can iteratively solve the optimization by alternatingly optimizing with respect to $(\W,\cc)$ and $\B$ while holding the other fixed. The idea of using auxiliary variable was also used in~\cite{BA_CVPR15} for learning binary codes, but~\cite{BA_CVPR15} only solves for case where hash function is linear autoencoder. %Furthermore, the way they solve for the binary constraint is also different from us, detailed in section~\ref{subsub:B_step_un}.

As mentioned in~\cite{DBLP:conf/nips/WeissTF08}, a good binary code  not only should have similarity preserving property but also should have independent and balancing properties. That is different bits are independent to each other and each bit has a $50\%$ chance of being $1$ or $-1$. So we add two more constraints (independence and balance) to problem~(\ref{eq:obj_2}). The new objective function is defined as
\begin{eqnarray}
\min_{\W,\cc,\B} J &=& \frac{1}{2m} \norm{\X-\W^{(n-1)}\B-\cc^{(n-1)}\1_{1\times m}}^2 \nonumber \\ 
{}&&+\frac{\lambda_1}{2}\sum_{l=1}^n \norm{\W^{(l)}}^2 \label{eq:obj_3}
\end{eqnarray}
\hspace{0.3cm} s.t. \\
\vspace{-0.5cm}
\begin{equation}
\B = sgn(\H^{(n-1)}) \label{eq:binary}
\end{equation} 
\begin{equation}
\frac{1}{m}\B\B^T = \I \label{eq:independent}
\end{equation}
\begin{equation}
\frac{1}{m}\norm{\B\1_{m\times 1}}^2 =0 \label{eq:balance}
\end{equation}
Where $\I$ is identity matrix. The problem~(\ref{eq:obj_3}) under the constraints is still NP hard and difficult to solve because of the discrete variable $\B$. One way to handle with this difficulty is by relaxing the constraint~(\ref{eq:binary}) as $\B = \H^{(n-1)}$. With this approach, this binary solution is achieved by first relaxing the binary codes to a continuous space and then post-processing, i.e. thresholding, the continuous solution. Most existing approach follow this relaxation such as Deep Hashing~\cite{Liong_2015_CVPR}, Semantic Hashing~\cite{DBLP:SeH}, Spectral Hashing~\cite{DBLP:conf/nips/WeissTF08}, AnchorGraph Hashing~\cite{DBLP:conf/icml/LiuWKC11}, Semi-Supervised Hashing~\cite{DBLP:journals/pami/WangKC12}, LDAHash~\cite{DBLP:journals/pami/StrechaBBF12},~\textit{etc}. This relaxation simplifies the original binary constraint problem and its solution is suboptimal, i.e., the binary codes resulting of thresholded continuous codes is not necessary same as codes resulting by directly solving the thresholding process in the optimization. 

In order to achieve binary codes of better quality, we should solve the binary constraint during the learning of the hash function. Inspired by the regularization methods~\cite{malick:hal-00389552}, we rewrite~(\ref{eq:obj_3}) and constraints~(\ref{eq:binary}),~(\ref{eq:independent}),~(\ref{eq:balance}) as
\begin{eqnarray}
\min_{\W,\cc,\B} J &=& \frac{1}{2m} \norm{\X-\W^{(n-1)}\B-\cc^{(n-1)}\1_{1\times m}}^2 \nonumber \\ 
{}&&\hspace{-2em}+\frac{\lambda_1}{2}\sum_{l=1}^n \norm{\W^{(l)}}^2 + \frac{\lambda_2}{2m}\norm{\H^{(n-1)}-\B}^2  \label{eq:obj_4}
\end{eqnarray}
\hspace{0.3cm} s.t. 
\vspace{-0.2cm}
\begin{equation}
\B \in \{-1,1\}^{L\times m} \label{eq:binary_H}
\end{equation}
\begin{equation}
\frac{1}{m}\H^{(n-1)}(\H^{(n-1)})^T = \I \label{eq:independent_H}
\end{equation}
\begin{equation}
\frac{1}{m}\norm{\H^{(n-1)}\1_{m\times 1}}^2 =0 \label{eq:balance_H}
\end{equation}

The third term in~(\ref{eq:obj_4}) is to minimize the discretization error between the continuous code $\H^{(n-1)}$ and the binary code $\B$. It is shown in~\cite{malick:hal-00389552} that with sufficiently large $\lambda_2$, minimizing~(\ref{eq:obj_4}) under constraint~(\ref{eq:binary_H}) becomes close to the minimizing~(\ref{eq:obj_3}) under constraint~(\ref{eq:binary}). When $\lambda_2$ is sufficiently large, the optimization process will result $\B \approx \H^{(n-1)}$. So we can rewrite constraints (\ref{eq:independent}), (\ref{eq:balance}) by constraints (\ref{eq:independent_H}), (\ref{eq:balance_H}).

The recent work SDH~\cite{Shen_2015_CVPR} on supervised hashing also used idea of regularization method~\cite{malick:hal-00389552}. However, their work focused on supervised hashing; their formulation is based on the assumption that the resulted codes is good for linear classification;  %linear hash function 
furthermore, they did not consider independent and balancing properties of codes. They are  different from our work, focusing on unsupervised hashing, no assumption on codes, using deep neural network as hash function and considering independent and balancing properties of codes. 

Instead of solving~(\ref{eq:obj_4}) under many constraints, using Lagrange multipliers approach, we solve similar following problem
\begin{eqnarray}
\min_{\W,\cc,\B} J &=& \frac{1}{2m} \norm{\X-\W^{(n-1)}\B-\cc^{(n-1)}\1_{1\times m}}^2 \nonumber \\ 
{}&&\hspace{-2em}+\frac{\lambda_1}{2}\sum_{l=1}^n \norm{\W^{(l)}}^2 + \frac{\lambda_2}{2m}\norm{\H^{(n-1)}-\B}^2 \nonumber \\
{}&&\hspace{-2em}+\frac{\lambda_3}{2}\norm{\frac{1}{m}\H^{(n-1)}(\H^{(n-1)})^T-\I}^2 \nonumber \\
{}&&\hspace{-2em}+\frac{\lambda_4}{2m}\norm{\H^{(n-1)}\1_{m\times 1}}^2 \label{eq:obj_5}
\end{eqnarray}
\hspace{0.3cm} s.t. 
\begin{equation}
\B \in \{-1,1\}^{L\times m} \label{eq:binary_H1}
\end{equation}
\subsection{Optimization}
\label{subsec:UDH_opt}
To solve~(\ref{eq:obj_5}) under constraint~(\ref{eq:binary_H1}), we alternating optimize over $(\W,\cc)$ and $\B$.
\subsubsection{$(\W,\cc)$ step}
\label{subsub:W_step_un}
When fixing $\B$, the problem becomes unconstrained optimization. We used $L-BFGS$~\cite{Liu89onthe,lbfgs2} optimizer with backpropagation for solving it. The gradient of objective function $J$ (\ref{eq:obj_5}) w.r.t. different parameters are computed as follows
\begin{eqnarray}
\frac{\partial J}{\partial \W^{(n-1)}} &=& \frac{-1}{m}(\X-\W^{(n-1)}\B-\cc^{(n-1)}\1_{1\times m})\B^T \nonumber\\
{} && + \lambda_1 \W^{(n-1)}
\end{eqnarray}
\begin{equation}
\frac{\partial J}{\partial \cc^{(n-1)}} = \frac{-1}{m}\left( (\X-\W^{(n-1)}\B)\1_{m\times 1})-m\cc^{(n-1)} \right)
\end{equation}
Let us define
\begin{eqnarray}
\Delta^{(n-1)} &=& \left[ \frac{\lambda_2}{m}\left( \H^{(n-1)}-\B \right) \right. \nonumber\\
{}&&\hspace{-4em}+\frac{2\lambda_3}{m}\left( \frac{1}{m}\H^{(n-1)}(\H^{(n-1)})^T - \I\right)\H^{(n-1)}\nonumber \\
 {}&& \left. \hspace{-4em}+\frac{\lambda_4}{m}\left( \H^{(n-1)}\1_{m\times m} \right) \right]\odot f^{(n-1)'}(\Z^{(n-1)})
\end{eqnarray}
\begin{equation}
\Delta^{(l)} = \left( (\W^{(l)})^T\Delta^{(l+1)} \right) \odot f^{(l)'}(\Z^{(l)}),\forall l = n-2,\cdots,2
\end{equation}
where $\odot$ denotes Hadamard product; $\Z^{(l)} = \W^{(l-1)}\H^{(l-1)} + \cc^{(l-1)}\1_{1\times m}$, $l=2,\cdots,n$\\
Then, $\forall l = n-2,\cdots,1$, we have
\begin{equation}
\frac{\partial J}{\partial \W^{(l)}} = \Delta^{(l+1)}(\H^{(l)})^T +\lambda_1\W^{(l)}
\end{equation}
\begin{equation}
\frac{\partial J}{\partial \cc^{(l)}} = \Delta^{(l+1)}\1_{m\times 1}
\end{equation}
\subsubsection{$\B$ step}
\label{subsub:B_step_un}
When fixing $(\W,\cc)$, we can rewrite problem~(\ref{eq:obj_5}) as 
\begin{eqnarray}
\min_{\B} J &=& \norm{\X-\W^{(n-1)}\B-\cc^{(n-1)}\1_{1\times m}}^2 \nonumber\\
{}&&+\lambda_2 \norm{\H^{(n-1)}-\B}^2
\end{eqnarray}
\hspace{0.7cm} s.t. 
\begin{equation}
\B \in \{-1,1\}^{L\times m} \label{eq:binary_H11}
\end{equation}
Solving $\B$ is challenging because of binary constraints on $\B$. Here we use recent proposed method \textit{discrete cyclic coordinate descent}~\cite{Shen_2015_CVPR}. The advantage of this method is if we fix $L-1$ rows of $\B$ and only solve for the remaining row, we can achieve a closed-form solution for that row. It means that we can iteratively solve $\B$ row by row.\\ %In the other words, we can learn one bit at a time. 
Let $\V = \X-\cc^{(n-1)}\1_{1\times m}$; $\Q = (\W^{(n-1)})^T\V+\lambda_2\H^{(n-1)}$. For $k=1,\cdots L$, let $\w_k$ be $k^{th}$ column of $\W^{(n-1)}$; $\W_1$ the matrix $\W$ excluding $\w_k$;
$\q_k$ be $k^{th}$ column of $\Q^T$; $\bb_k^T$ be $k^{th}$ row of $\B$; $\B_1$ the matrix of $\B$ excluding $\bb_k^T$. We have closed-form for $\bb_k^T$ as
\begin{equation}
\bb_k^T = sgn(\q^T - \w_k^T\W_1\B_1)
\end{equation}

The proposed UDH-DNN method is summarized in Algorithm~\ref{alg1}. In the Algorithm~\ref{alg1}, $\B_{(t)}$ and $(\W,\cc)_{(t)}$ are values of $\B$ and $\{\W^{(l)},\cc^{(l)}\}_{l=1}^{n-1}$ at iteration $t$.

\begin{algorithm}[!t]
	\footnotesize
	\caption{Unsupervised Discrete Hashing with Deep Neural Network (UDH-DNN)}
	\begin{algorithmic}[1] 
		\Require 
			\Statex $\X = \{\x_i\}_{i=1}^{m} \in \R^{D\times m}$: training data; $L$: code length; $max\_iter$: maximum iteration number; $n$: number of layers; $\{s_l\}_{l=2}^{n}$: number of units of layers $2 \to n$ (Note: number of units of layers $n-1$ and $n$ should equal to $L$ and $D$, respectively.); $\lambda_1, \lambda_2, \lambda_3, \lambda_4$.
		\Ensure 
			\Statex Binary code $\B \in \R^{L\times m}$ of training data $\X$; parameters $\{\W^{(l)},\cc^{(l)}\}_{l=1}^{n-1}$
			\Statex 
			\State Initialize $\B_{(0)}$ using ITQ~\cite{DBLP:conf/cvpr/GongL11}
			\State %$\forall l = 1,\cdots,n-1$, initialize $\cc^{(l)} = \mathbf{0}_{s_{l+1}\times 1}$. Initialize $\W^{(1)}$ by getting the top $s_2$ eigenvectors from the covariance matrix of $\X$. $\forall l = 2,\cdots,n-2$, initialize $\W^{(l)}$ by getting the top $s_{l+1}$ eigenvectors from the covariance matrix of $\H^{(l)}$. 
			Initialize $\{\cc^{(l)}\}_{l=1}^{n-1} = \mathbf{0}_{s_{l+1}\times 1}$. Initialize $\W^{(1)}$ by getting the top $s_2$ eigenvectors from the covariance matrix of $\X$. Initialize $\{\W^{(l)}\}_{l=2}^{n-2}$ by getting the top $s_{l+1}$ eigenvectors from the covariance matrix of $\H^{(l)}$. Initialize $\W^{(n-1)} = \I_{D\times L}$
			\State Compute $(\W,\cc)_{(0)}$ with $(\W,\cc)$ step (Sec.~\ref{subsub:W_step_un}), using $\B_{(0)}$ as fixed value and using initialized $\{\W^{(l)},\cc^{(l)}\}_{l=1}^{n-1}$ (at line 2) as starting point for $L-BFGS$.
			\For{$t = 1 \to max\_iter$}
				\State Compute $\B_{(t)}$ by iteratively learning row by row with $\B$ step (Sec.~\ref{subsub:B_step_un}), using $(\W,\cc)_{(t-1)}$ as fixed values.
				\State Compute $(\W,\cc)_{(t)}$ with $(\W, \cc)$ step (Sec.~\ref{subsub:W_step_un}), using $\B_{(t)}$ as fixed value and using $(\W,\cc)_{(t-1)}$ as starting point for $L-BFGS$.
			\EndFor
			\State Return $\B_{(max\_iter)}$ and $(\W,\cc)_{(max\_iter)}$
    \end{algorithmic}
    \label{alg1}
\end{algorithm}
\section{Evaluation of Unsupervised Discrete Hashing with Deep Neural Network}
\label{sec:eva_udh-dnn}
This section presents results of UDH-DNN. We compare UDH-DNN with following state-of-the-art unsupervised hashing methods: Spectral Hashing (SH)~\cite{DBLP:conf/nips/WeissTF08}, 
%Semantic Hashing (SeH)~\cite{DBLP:SeH}, AnchorGraph Hashing (AGH)~\cite{DBLP:conf/icml/LiuWKC11},
 Iterative Quantization (ITQ)~\cite{DBLP:conf/cvpr/GongL11}, Binary Autoencoder (BA)~\cite{BA_CVPR15}, Spherical Hashing (SPH)~\cite{CVPR12:SphericalHashing}, K-means Hashing (KMH)~\cite{DBLP:conf/cvpr/HeWS13}. For all compared methods, we use the codes and the suggested parameters provided by the authors.
\subsection{Dataset, implementation note, and evaluation protocol}
\label{subsec:data-imp-eva}
\paragraph{CIFAR-10}  CIFAR-10~\cite{Krizhevsky09} contains 60,000 color images of 10 classes. Each image has size of $32\times 32$. The training set contains 50,000 images, and the testing set contains 10,000 images. In this experiment, we ignore the class labels. As standardly done in the literature~\cite{DBLP:conf/cvpr/GongL11,BA_CVPR15}, we extract 320-$D$ GIST features~\cite{gist} from each image.
%\paragraph{NUS-WIDE} NUS-WIDE~\cite{nus-wide-civr09} dataset contains 269,648 images collected from Flickr. The training set contains 161,789 images and the testing set contains 107,859 images. The provided 128-dimensional wavelet features features are used. 
\paragraph{MNIST} The MNIST~\cite{mnistlecun} dataset consists of 70,000 handwritten digit
images of 10 classes (labeled from 0 to 9). Each image has size of $28\times28$. The training set contains 60,000 samples, and the test set contains 10,000 samples. In this experiment, we ignore the class labels. Each image was represented as a 784-$D$ gray-scale feature vector by using its intensity. 
\paragraph{SIFT1M} SIFT1M~\cite{herve_pami2011} dataset contains 128-$D$ SIFT vectors. This is standard dataset used for evaluating large scale approximate nearest neighbor search.  There are 1M vectors for indexing; 100K vectors for training (separated from indexing set) and 10K vectors for testing. 

\paragraph{Implementation note}
In our deep model, we use $n=5$ layers (including input and output layer). The activation functions for layers $2$ and $3$ are sigmoid functions; for layers $4$ and $5$ are linear functions. The parameters $\lambda_1$, $\lambda_2$, $\lambda_3$ and $\lambda_4$ were empirically set as $10^{-5}$, $5\times 10^{-2}$, $10^{-2}$ and $10^{-6}$, respectively. The max iteration number $max\_iter$ is set to 10. 

For the CIFAR-10 and MNIST datasets, the number of units in hidden layers $2,3,4$ were empirically set as $[90 \to 20 \to 8]$, $[90 \to 30 \to 16]$, $[120 \to 50 \to 32]$  and $[160 \to 110 \to 64]$ for the 8, 16, 32 and 64 bits respectively. 
For the SIFT1M dataset, the number of units in hidden layers $2,3,4$ were empirically set as 
$[90 \to 20 \to 8]$, $[90 \to 30 \to 16]$, $[100 \to 50 \to 32]$  and $[100 \to 80 \to 64]$ for the 8, 16, 32 and 64 bits respectively.

\paragraph{Evaluation metric} 
We follow standard setting widely used in unsupervised hashing~\cite{DBLP:conf/cvpr/GongL11,CVPR12:SphericalHashing,DBLP:conf/cvpr/HeWS13,BA_CVPR15} using Euclidean nearest neighbors to create ground truths for queries. Number of ground truths are set as in~\cite{BA_CVPR15}. For datasets CIFAR-10 and MNIST, for each query, we use $50$ its Euclidean nearest neighbors as ground truth. For large scale dataset SIFT1M, for each query, we use $10,000$ its Euclidean nearest neighbors as ground truth.

We used the following evaluation metrics~\cite{DBLP:conf/cvpr/GongL11,BA_CVPR15} to measure the performance of methods. 1) mean average precision (mAP) which not only considers precision but also considers rank of retrieval results; %2) precision when $K$ nearest neighbors are returned (precision$@K$); 3) precision of Hamming radius $r$ (precision$@r$) which measure precision on retrieved images having Hamming distance to query $\le r$ (if no images satisfy, we report zero precision). 
 2) precision of Hamming radius $r$ (precision$@r$) which measure precision on retrieved images having Hamming distance to query $\le r$ (if no images satisfy, we report zero precision).  

\subsection{Retrieval results}
\subsubsection{Results on CIFAR-10 dataset} 
\begin{figure*}[!t]
\centering
\subfigure[]{
       \includegraphics[scale=0.35]{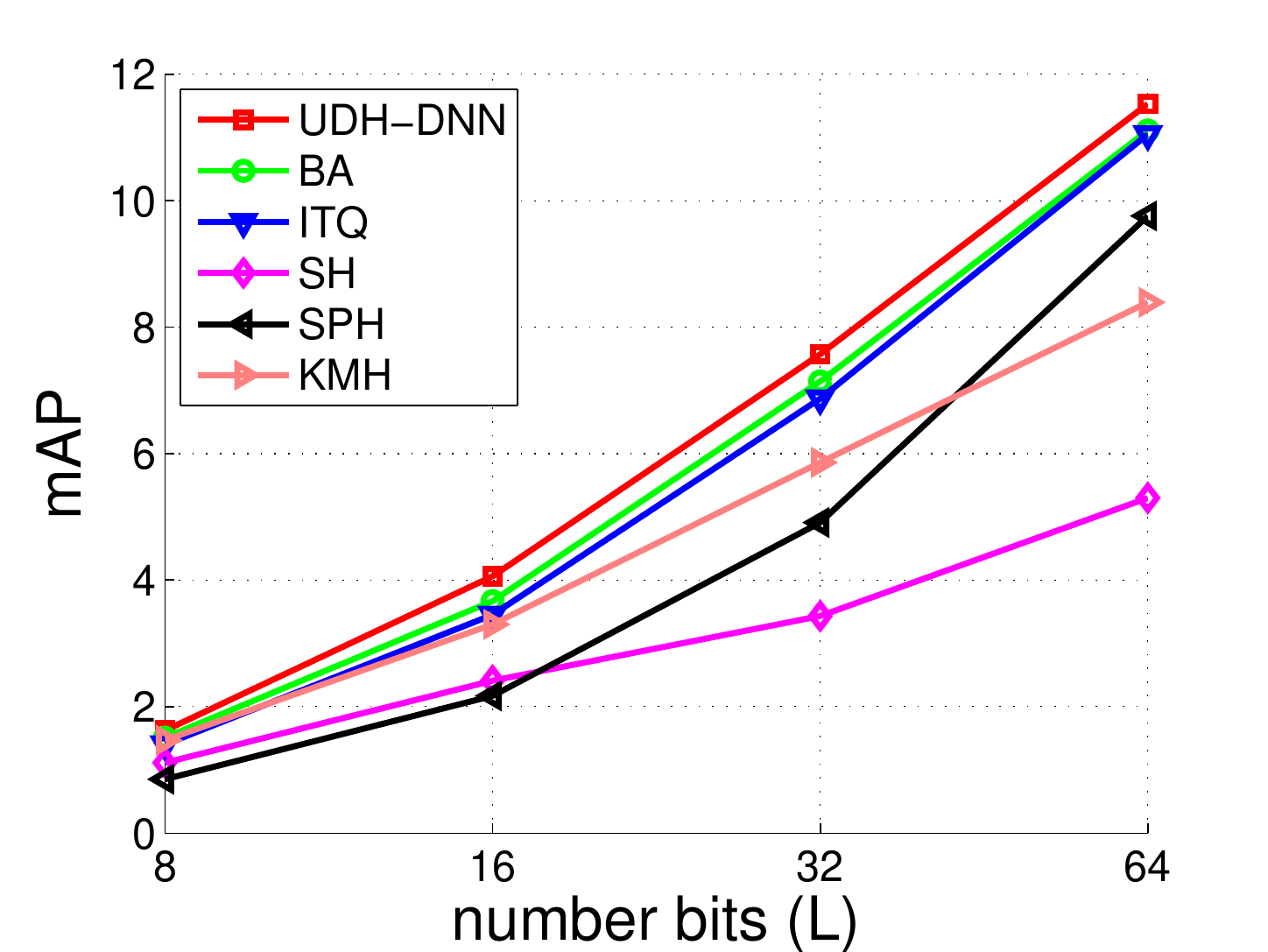}
       \label{fig:cifar_mAP}
}
%\subfigure[]{
%       \includegraphics[scale=0.35]{cifar10_pK.pdf}
%       \label{fig:cifar_pK}
%}
\subfigure[]{
       \includegraphics[scale=0.35]{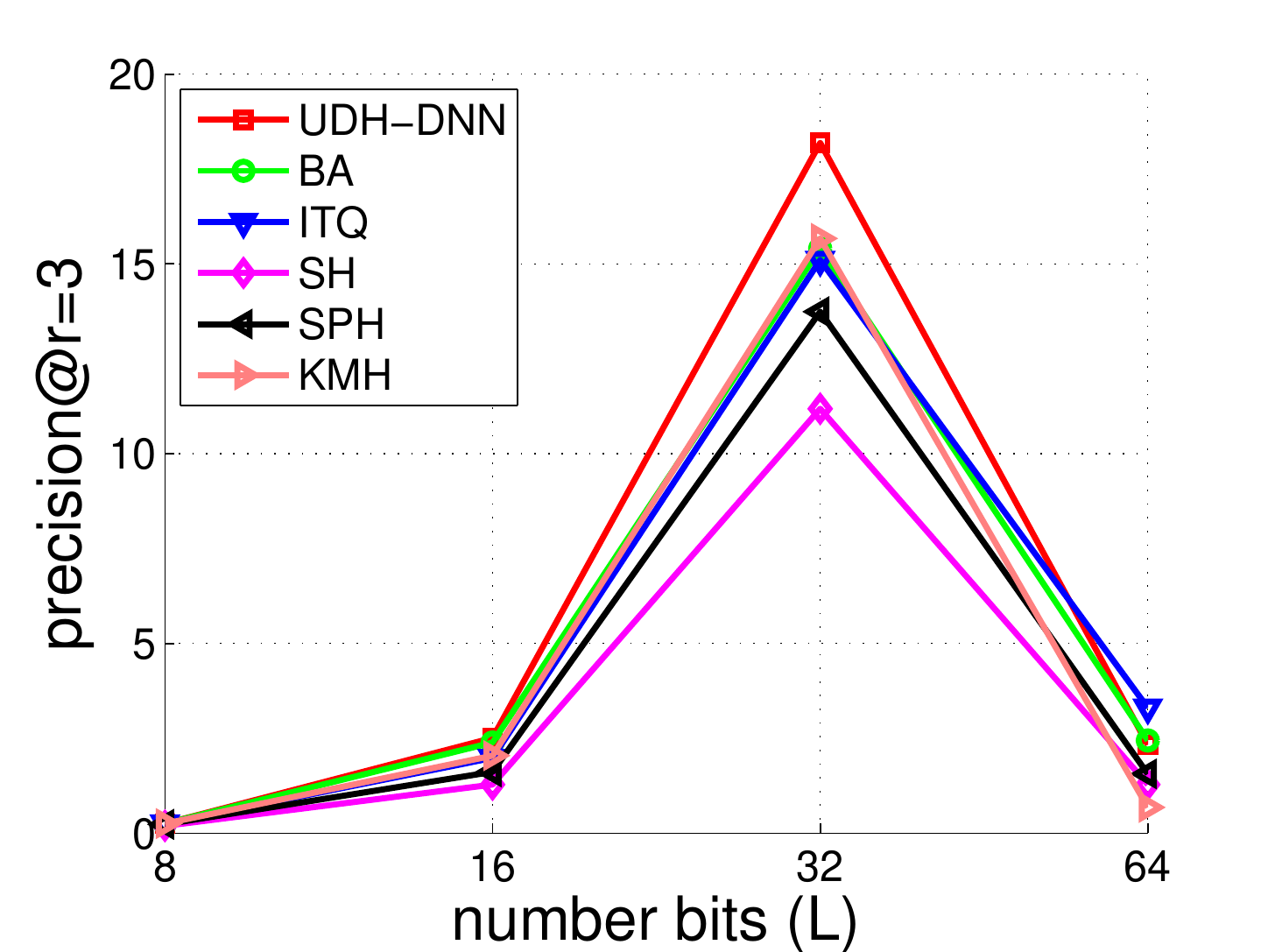} 
       \label{fig:cifar_pr3}
}
\subfigure[]{
       \includegraphics[scale=0.35]{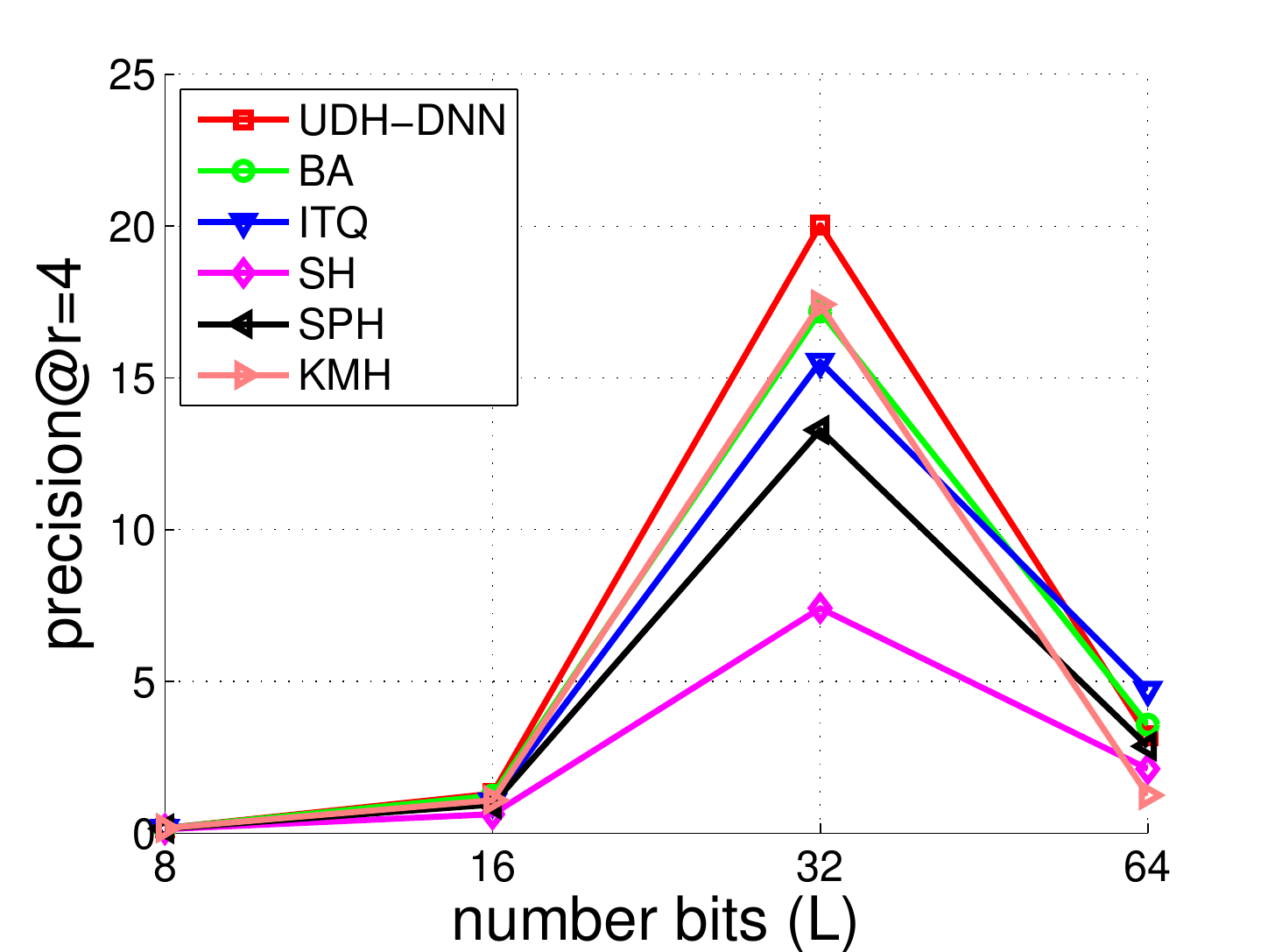} 
       \label{fig:cifar_pr4}
}
%\caption[]{Comparative evaluation on CIFAR-10 dataset. \ref{fig:cifar_pK} and ~\ref{fig:cifar_mAP}: Precision and mAP when considering top 50 returned nearest neighbors.~\ref{fig:cifar_pr}: Precision when considering images with in Hamming distance 2.}
\caption[]{Comparative evaluation on CIFAR-10 dataset.~\ref{fig:cifar_mAP}: mAP.   \ref{fig:cifar_pr3} and \ref{fig:cifar_pr4}: Precision when considering retrieved images with in Hamming distance 3 and 4, respectively. Number of ground truths for each query = 50.}
\label{fig:cifar10}
\end{figure*}

Figure~\ref{fig:cifar10} shows retrieval results of different methods with different code lengths $L$ on CIFAR-10 dataset. 

In term of mAP, the proposed UDH-DNN achieves the best results for all code lengths. The improvement is more clear at high $L$. The mAP of UDH-DNN consistent outperforms binary autoencoder (BA)~\cite{BA_CVPR15} which is current state-of-the-art unsupervised hashing method. 

When precision of Hamming radius $r$ is used, the following observations are consistent for both $r=3$ and $r=4$. The UDH-DNN is comparable to other methods at low $L$ (i.e. $L = 8, 16$). At $L = 32$, UDH-DNN significant outperforms other methods. When $L = 64$, all methods decrease the precision. The reason is that many query images have no neighbors at a Hamming distance of $r$ or less and we report zero precision for those cases. The precision of UDH-DNN is lower than some compared methods at $L=64$. However, we note a larger variance: the highest precision is achieved by UDH-DNN at $L=32$ for both $r=3$ and $r=4$ cases.

\paragraph{Comparison with Deep Hashing (DH)~\cite{Liong_2015_CVPR}}
We also compare our UDH-DNN with the Deep Hashing (DH)~\cite{Liong_2015_CVPR}. Because the implementation of DH is not available, we set up our experiments similar to ~\cite{Liong_2015_CVPR} to make a fair comparison. We randomly sample 1,000 images, 100 per class, as testing set; the remaining 59,000 images are used as training set. Each image is represented by 512-$D$ GIST descriptor~\cite{gist}. The ground truths of queries are based on their class labels\footnote{It is worth noting that in the evaluation of unsupervised hashing, instead of using class label as ground truths, most state-of-the-art methods~\cite{DBLP:conf/cvpr/GongL11,CVPR12:SphericalHashing,DBLP:conf/cvpr/HeWS13,BA_CVPR15} use Euclidean nearest neighbors as ground truths for queries.}. Similar to~\cite{Liong_2015_CVPR}, we report comparative results in term of mAP at code lengths $L=16, 32, 64$ and the precision at Hamming radius of $r=2$ at code lengths $L=16,32$. We perform the experiments 10 times and report the average performance. 
The comparative results are presented in the Table~\ref{tab:compare_DH_cifar512}.
\begin{table}[!t]
\footnotesize
\centering
\caption{Comparison with Deep Hashing (DH)~\cite{Liong_2015_CVPR} at different code lengths on the CIFAR-10 dataset. The results of DH are obtained from corresponding paper.} 
\label{tab:compare_DH_cifar512}
\begin{tabular}{|c|c|c|c|c|c|}
\hline
Method					&\multicolumn{3}{|c|}{mAP} 	 &\multicolumn{2}{|c|}{Precision$@r=2$}\\ 
							&$L=16$	&$L=32$	&$L=64$			&$L=16$	&$L=32$\\\hline
DH~\cite{Liong_2015_CVPR}   &16.17  &16.62  &16.96          &23.33  &15.77 \\
UDH-DNN	 &~\textbf{16.83}  &\textbf{17.52}  &\textbf{18.02}	&\textbf{24.97}  &\textbf{22.20} \\\hline 
\end{tabular}
\end{table} 
It is clearly showed in Table~\ref{tab:compare_DH_cifar512} that the proposed UDH-DNN outperforms DH~\cite{Liong_2015_CVPR} at all code lengths, in both mAP and precision of Hamming radius. It is because the UDH-DNN %not only considers the independent, balancing properties but also the similarity preserving property of codes. 
contains all necessary criteria for producing good binary codes. 
Furthermore, instead of doing the relaxation on the binary constraint when learning the network as DH~\cite{Liong_2015_CVPR}, we directly solve the binary constraint during the learning process. 

\subsubsection{Results on MNIST dataset} 
\begin{figure*}[!t]
\centering
\subfigure[]{
       \includegraphics[scale=0.35]{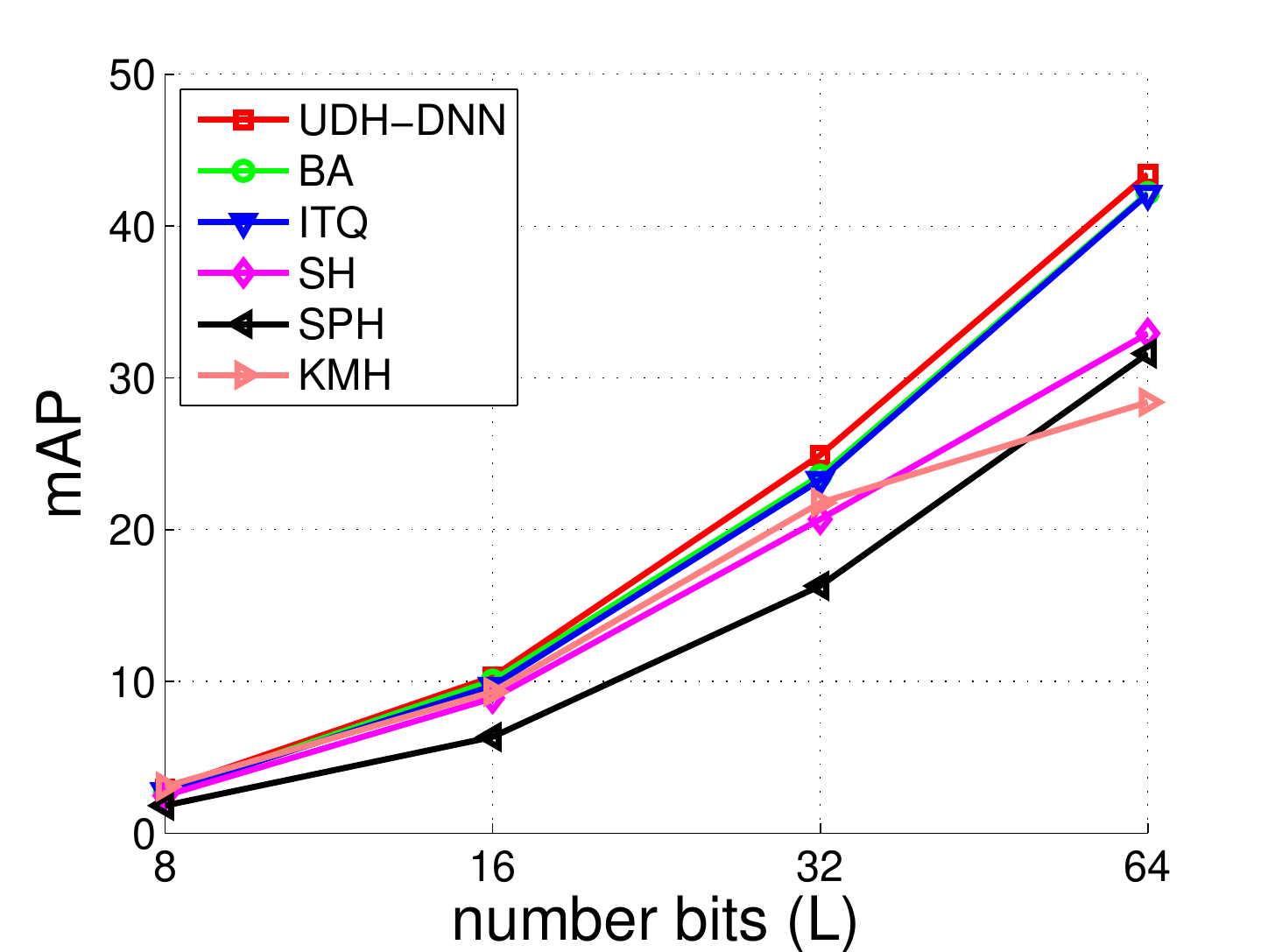}
       \label{fig:mnist_mAP}
}
%\subfigure[]{
%       \includegraphics[scale=0.35]{mnist_pK.pdf}
%       \label{fig:mnist_pK}
%}
\subfigure[]{
       \includegraphics[scale=0.35]{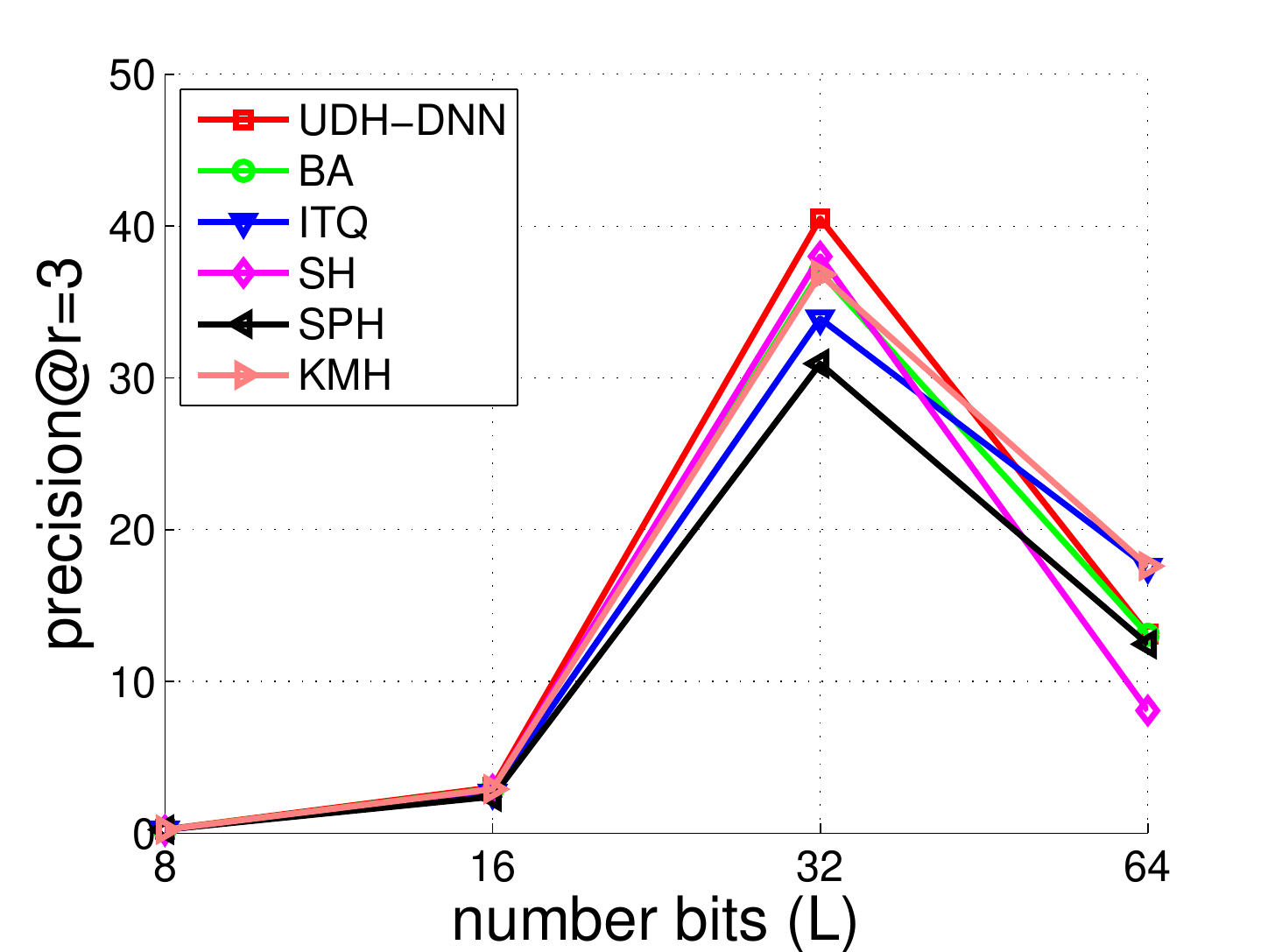} 
       \label{fig:mnist_pr3}
}
\subfigure[]{
       \includegraphics[scale=0.35]{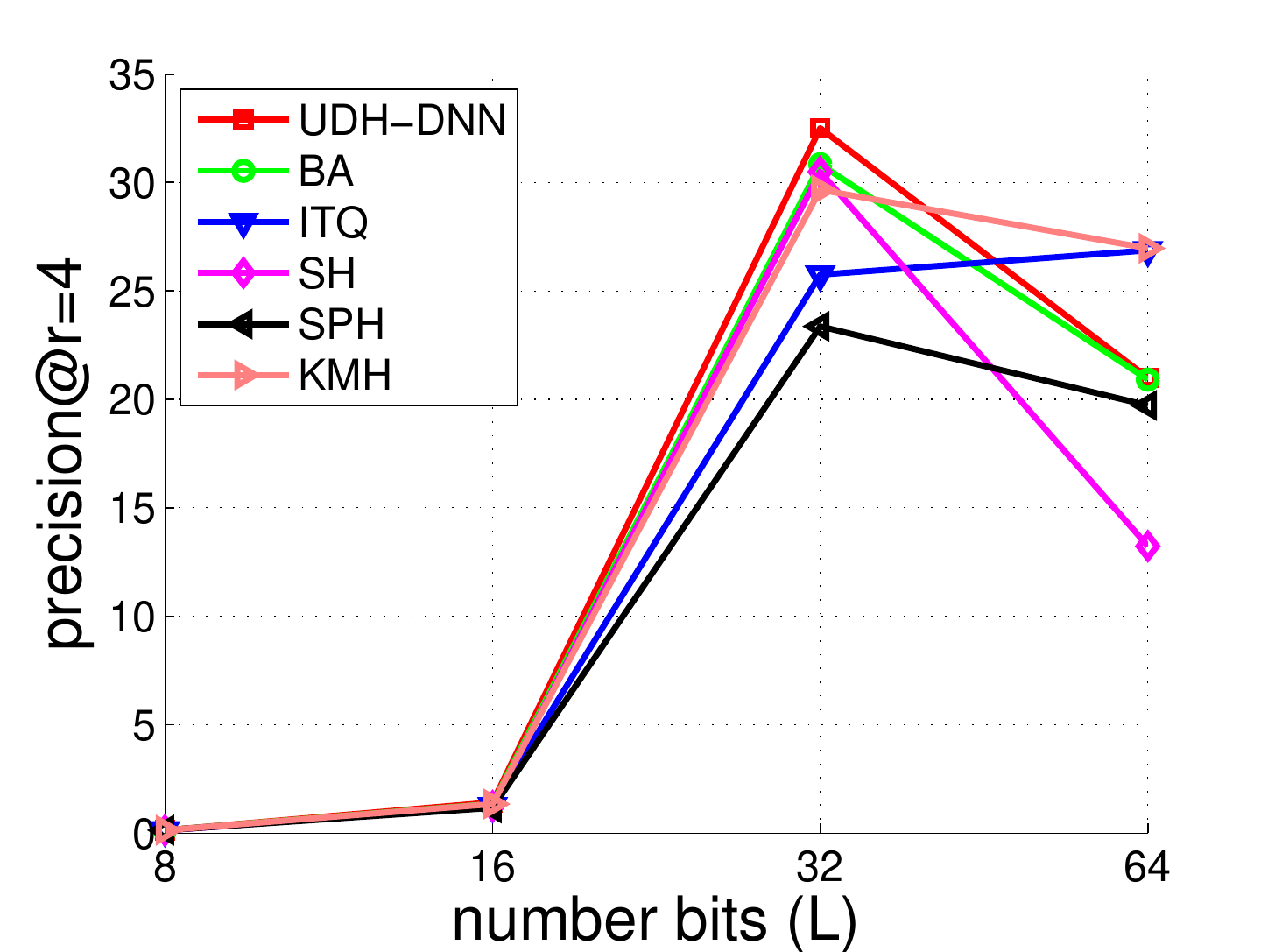} 
       \label{fig:mnist_pr4}
}
\caption[]{Comparative evaluation on MNIST dataset.~\ref{fig:cifar_mAP}: mAP.   \ref{fig:cifar_pr3} and \ref{fig:cifar_pr4}: Precision when considering retrieved images with in Hamming distance 3 and 4, respectively. Number of ground truths for each query = 50.}
\label{fig:mnist}
\end{figure*}

Figure~\ref{fig:mnist} shows retrieval results of different methods with different code lengths $L$ on MNIST dataset. 

The results are quite consistent with the results on the CIFAR-10 dataset. The proposed UDH-DNN achieves the best mAP for all code lengths. The mAP improvement is more clear at high $L$.

When precision of Hamming radius $r$ is used, all methods achieve similar precision at low $L$ ($L=8, 16$). At $L=32$, UDH-DNN outperforms other methods by a fair margin.
For large $L$, i.e. $L = 64$, except for ITQ which slightly increase precision when $r=4$, all methods decrease the precision. The precision of UDH-DNN is lower than some compared methods at $L=64$. However, it is worth noting that the highest precision is achieved by UDH-DNN (at $L=32$).% for both $r=3$ and $r=4$ cases.

\subsubsection{Results on SIFT1M dataset} 
\begin{figure*}[!t]
\centering
\subfigure[]{
       \includegraphics[scale=0.35]{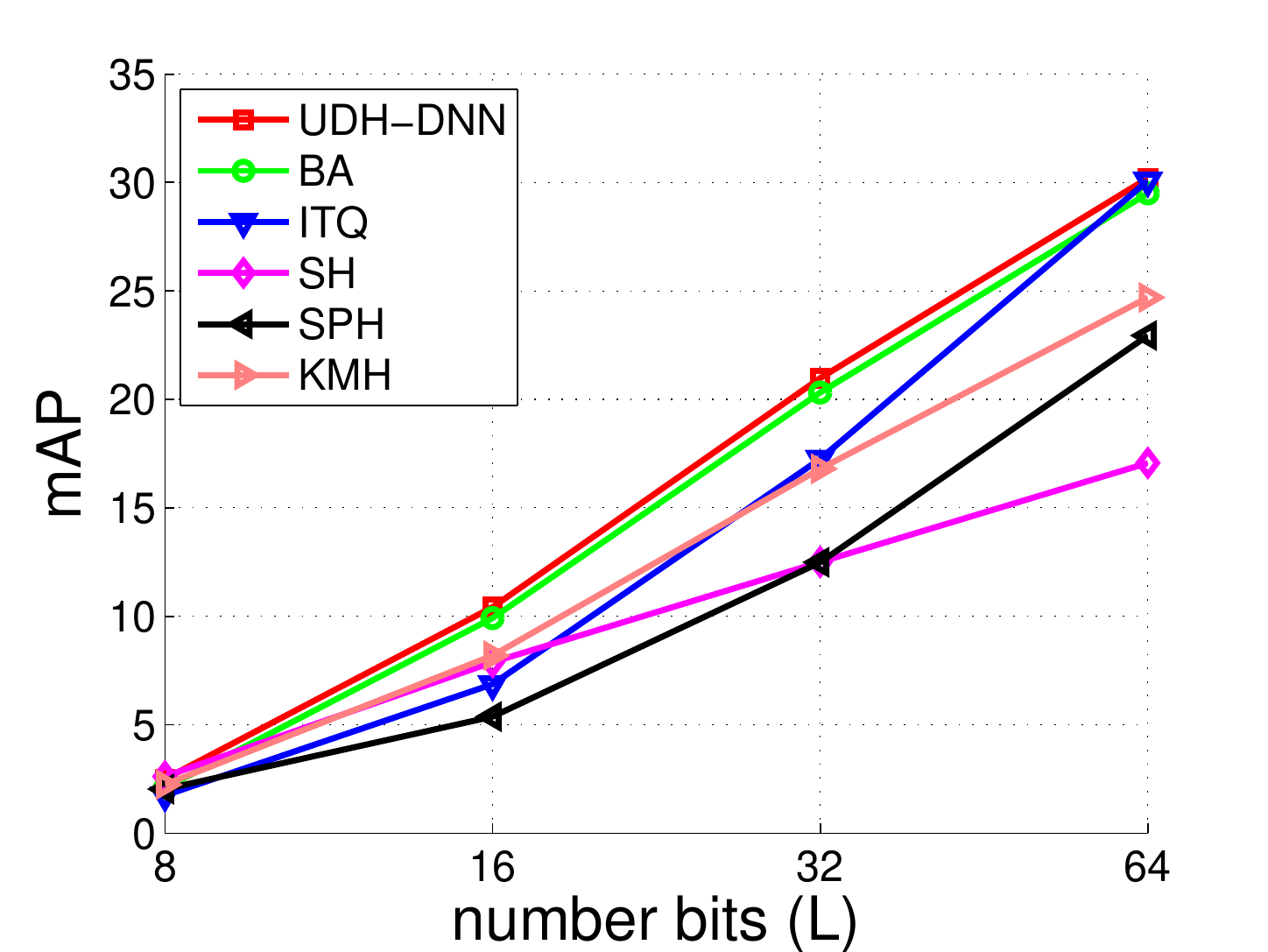}
       \label{fig:sift1m_mAP}
}
\subfigure[]{
       \includegraphics[scale=0.35]{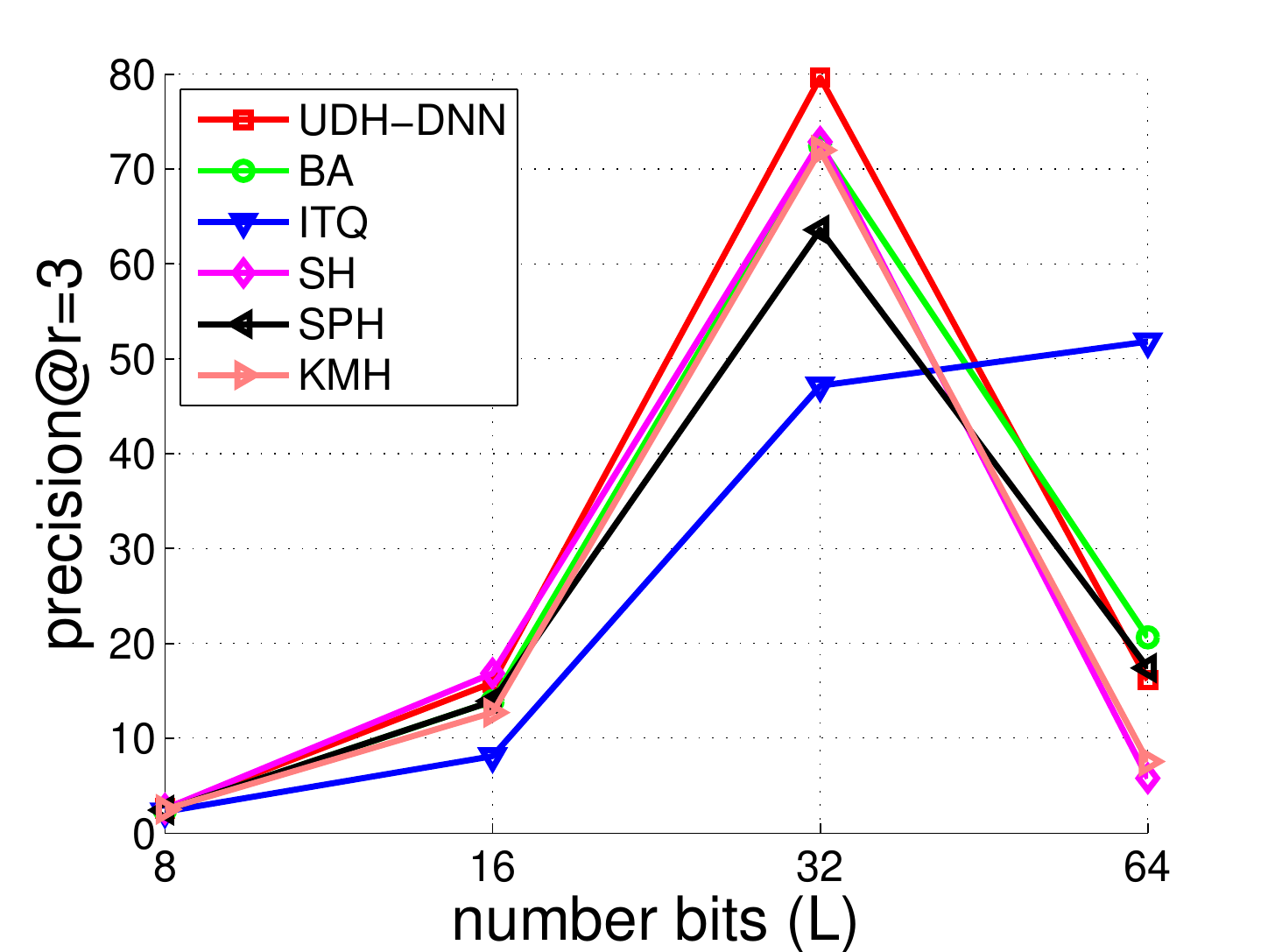} 
       \label{fig:sift1m_pr3}
}
\subfigure[]{
       \includegraphics[scale=0.35]{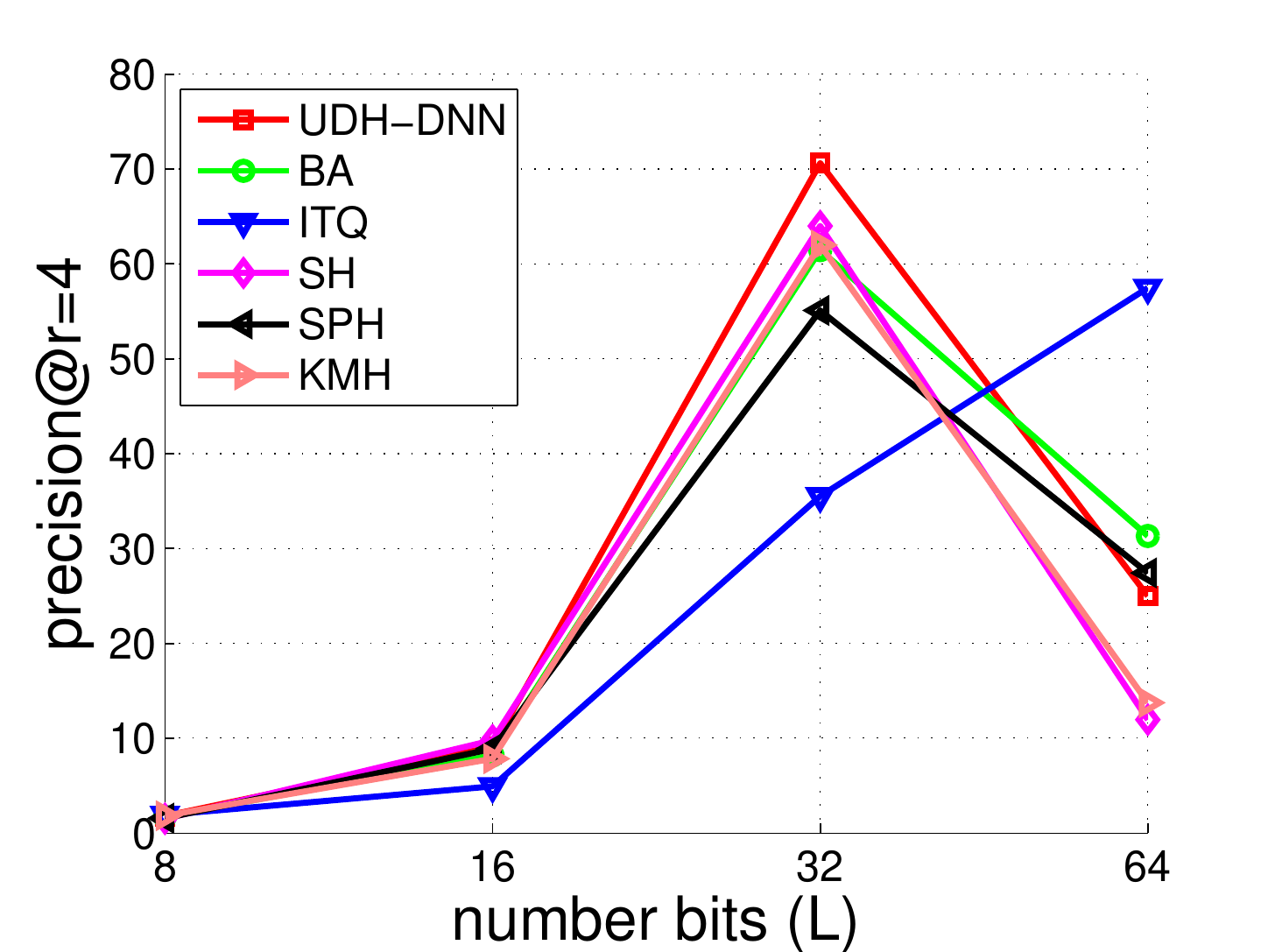} 
       \label{fig:sift1m_pr4}
}
\caption[]{Comparative evaluation on SIFT1M dataset.~\ref{fig:sift1m_mAP}: mAP.   \ref{fig:sift1m_pr3} and \ref{fig:sift1m_pr4}: Precision when considering retrieved images with in Hamming distance 3 and 4, respectively. Number of ground truths for each query = 10,000.}
\label{fig:sift1m}
\end{figure*}
As computing mAP is slow on this large dataset, we consider top-$10,000$ returned neighbors when computing mAP. Figure~\ref{fig:sift1m} shows retrieval results of different methods with different code lengths $L$ on SIFT1M dataset. 

In term of mAP, the proposed UDH-DNN is outperform all compared methods. It is slightly better than the current state-of-the-art unsupervised hashing binary autoencoder (BA)~\cite{BA_CVPR15}. 

In term of precision of Hamming radius, the results of UDH-DNN are consistent to its results on CIFAR-10 and MNIST. All methods achieve similar precision at low $L$ ($L=8, 16$). At $L=64$, precision of UDH-DNN is lower than some methods. However, the highest precision is achieved by UDH-DNN at $L=32$ and it is much better than the competitors.
\section{Supervised Discrete Hashing with Deep Neural Network (SDH-DNN)}
\label{sec:sdh-dnn}
There are several approaches proposed to leverage the label information when learning binary codes in the supervised hashing. In~\cite{DBLP:journals/pami/StrechaBBF12,minhdo_hash2014}, binary codes are learned such that they minimize Hamming distance between samples belonging to same class, while maximizing the Hamming distance between samples belonging to different classes. In~\cite{Shen_2015_CVPR}, the binary codes are learned such that they minimize the $l_2$ loss w.r.t. the ground truth labels. 

In this work, we adapt the approach proposed in kernel-based supervised hashing (KSH)~\cite{CVPR12:Hashing} to leverage the label information. The main idea is to learn binary codes such that the Hamming distance between binary codes of samples are high correlated with the pre-computed pairwise label matrix. In the other words, the binary codes should preserve the semantic (label) similarity between samples. It worth noting that in KSH~\cite{CVPR12:Hashing} the hash functions are linear and are defined in kernel space of inputs. The independent, balancing criteria are not considered in KSH~\cite{CVPR12:Hashing}. 

In general, the network structure of SDH-DNN is similar to the proposed UDH-DNN, excepting that the last layer preserving reconstruction is removed. The layer $n-1$ in UDH-DNN will become the last layer in SDH-DNN. The semantic preservation property in SDH-DNN is constrained on output of its last layer. 

\subsection{Formulation of SDH-DNN}
Following KSH~\cite{CVPR12:Hashing}, we fist define the pairwise label matrix $\S$ as
\begin{equation}
\S_{ij} = \left\{ \begin{array}{ll}
1 & \textrm{if $\x_i$ and $\x_j$ are same class}\\
-1 & \textrm{if $\x_i$ and $\x_j$ are not same class}
\end{array} \right.
\label{eq:S}
\end{equation}
The goal of learning process is to learn hash function which generating discriminative codes such that similar pairs can be perfectly distinguished from dissimilar pair by using Hamming distance in the code space. In the other words, the Hamming distance between learned binary codes should correlate with the matrix $\S$. Formally, the binary codes $B$ should satisfy
\begin{equation}
\min_{\B\in \{-1,1\}^{L\times m}} \mathcal{Q} = \norm{\frac{1}{L} \B^T\B - \S}^2
\end{equation}
Using the idea of regularization as the unsupervised hashing (Sec.~\ref{sec:udh-dnn}), we integrate the above criterion to our model by solving the following constrained optimization
\begin{eqnarray}
\min_{\W,\cc,\B} J &=& \frac{1}{2m}\norm{\frac{1}{L} (\H^{(n)})^T\H^{(n)} - \S}^2 \nonumber\\
{}&&\hspace{-4em}+\frac{\lambda_1}{2}\sum_{l=1}^n \norm{\W^{(l)}}^2 + \frac{\lambda_2}{2m}\norm{\H^{(n)}-\B}^2 \nonumber \\
{}&&\hspace{-4em}+\frac{\lambda_3}{2}\norm{\frac{1}{m}\H^{(n)}(\H^{(n)})^T-\I}^2 +\frac{\lambda_4}{2m}\norm{\H^{(n)}\1_{m\times 1}}^2 \nonumber \\
 \label{eq:obj_sup2}
\end{eqnarray}
\hspace{0.3cm}s.t. 
\begin{equation}
\B \in \{-1,1\}^{L\times m} \label{eq:binary_H1_sup2}
\end{equation}
The main difference in formulation between the proposed UDH-DNN~(\ref{eq:obj_5}) and the proposed SDH-DNN~(\ref{eq:obj_sup2}) is that the reconstruction term which indirectly preserves the neighbor similarity in UDH-DNN~(\ref{eq:obj_5}) is replaced by the term preserving the semantic (label) similarity in SDH-DNN~(\ref{eq:obj_sup2}).

\subsection{Optimization}
To solve~(\ref{eq:obj_sup2}) under constraint~(\ref{eq:binary_H1_sup2}), we alternating optimize over $(\W,\cc)$ and $\B$.
\subsubsection{$(\W,\cc)$ step}
\label{subsub:W_step_sup}
When fixing $\B$, (\ref{eq:obj_sup2}) becomes unconstrained optimization. We used $L-BFGS$~\cite{Liu89onthe} optimizer with backpropagation for solving it. The gradient of objective function $J$ w.r.t. different parameters are computed as follows.

Let
\begin{eqnarray}
\Delta^{(n)} &=& \nonumber \\
{}&&\hspace{-4em}\left[ \frac{1}{mL}\H^{(n)}\left( \V+\V^T \right)+\frac{\lambda_2}{m}\left( \H^{(n)}-\B \right) \right.  \nonumber\\
{}&&\hspace{-4em}+\frac{2\lambda_3}{m}\left( \frac{1}{m}\H^{(n)}(\H^{(n)})^T - \I\right)\H^{(n)}\nonumber \\
 {}&& \left. \hspace{-4em}+\frac{\lambda_4}{m}\left( \H^{(n)}\1_{m\times m} \right) \right]\odot f^{(n)'}(\Z^{(n)})
\end{eqnarray}
where $\V = \frac{1}{L}(\H^{(n)})^T\H^{(n)} - \S$. \\
Let
\begin{equation}
\Delta^{(l)} = \left( (\W^{(l)})^T\Delta^{(l+1)} \right) \odot f^{(l)'}(\Z^{(l)}),\forall l = n-1,\cdots,2
\end{equation}
where $\odot$ denotes Hadamard product; $\Z^{(l)} = \W^{(l-1)}\H^{(l-1)} + \cc^{(l-1)} \1_{1\times m}$, $l=2,\cdots,n$.\\
$\forall l = n-1,\cdots,1$, we have
\begin{equation}
\frac{\partial J}{\partial \W^{(l)}} = \Delta^{(l+1)}(\H^{(l)})^T +\lambda_1\W^{(l)}
\end{equation}
\begin{equation}
\frac{\partial J}{\partial \cc^{(l)}} = \Delta^{(l+1)}\1_{m\times 1}
\end{equation}

\subsubsection{$\B$ step}
\label{subsub:B_step_sup}
When fixing $(\W,\cc)$, we can rewrite problem~(\ref{eq:obj_sup2}) as 
\begin{equation}
\min_{\B} J =  \norm{\H^{(n)}-\B}^2
\label{eq:obj_sup3}
\end{equation}
s.t. 
\begin{equation}
\B \in \{-1,1\}^{L\times m} \label{eq:binary_H3}
\end{equation}
It is easy to see that the solution for~(\ref{eq:obj_sup3}) under constraint~(\ref{eq:binary_H3}) is $\B = sgn(\H^{(n)})$.

The proposed SDH-DNN method is summarized in Algorithm~\ref{alg2}. In the Algorithm~\ref{alg2}, $\B_{(t)}$ and $(\W,\cc)_{(t)}$ are values of $\B$ and $\{\W^{(l)},\cc^{(l)}\}_{l=1}^{n-1}$ at iteration $t$.

\begin{algorithm}[!t]
	\footnotesize
	\caption{Supervised Discrete Hashing with Deep Neural Network (SDH-DNN)}
	\begin{algorithmic}[1] 
		\Require 
			\Statex $\X = \{\x_i\}_{i=1}^{m} \in \R^{D\times m}$: training data; $\Y \in R^{m\times 1}$: training label vector; $L$: code length; $max\_iter$: maximum iteration number; $n$: number of layers; $\{s_l\}_{l=2}^{n}$: number of units of layers $2 \to n$ (Note: number of units of layer $n$ should equal to $L$); $n_s$: number of samples per class for computing pairwise label matrix $\S$; $\lambda_1, \lambda_2, \lambda_3, \lambda_4$.
		\Ensure 
			\Statex Binary code $\B \in \R^{L\times m}$ of training data $\X$ and parameters $\{\W^{(l)},\cc^{(l)}\}_{l=1}^{n-1}$
			\Statex 
			\State Random select $n_s$ samples per class and compute pairwise label matrix $\S$ using~(\ref{eq:S}).
			\State Initialize $\B_{(0)}$ using ITQ~\cite{DBLP:conf/cvpr/GongL11}
			\State Initialize $\{\cc^{(l)}\}_{l=1}^{n-1} = \mathbf{0}_{s_{l+1}\times 1}$. Initialize $\W^{(1)}$ by getting the top $s_2$ eigenvectors from the covariance matrix of $\X$. Initialize $\{\W^{(l)}\}_{l=2}^{n-1}$ by getting the top $s_{l+1}$ eigenvectors from the covariance matrix of $\H^{(l)}$. 
			\State Compute $(\W,\cc)_{(0)}$ with $(\W,\cc)$ step (Sec.~\ref{subsub:W_step_sup}), using $\B_{(0)}$ as fixed values and using initialized $\{\W^{(l)},\cc^{(l)}\}_{l=1}^{n-1}$ (at line 3) as starting point for $L-BFGS$.
			\For{$t = 1 \to max\_iter$}
				\State Compute $\B_{(t)}$ with $\B$ step (Sec.~\ref{subsub:B_step_sup}), using $(\W,\cc)_{(t-1)}$ as fixed values.
				\State Compute $(\W,\cc)_{(t)}$ with $(\W, \cc)$ step (Sec.~\ref{subsub:W_step_sup}), using $\B_{(t)}$ as fixed values and using $(\W,\cc)_{(t-1)}$ as starting point for $L-BFGS$.
			\EndFor
			\State Return $\B_{(max\_iter)}$ and $(\W,\cc)_{(max\_iter)}$
    \end{algorithmic}
    \label{alg2}
\end{algorithm}

%\subsection{Leveraging label information via both $l_2$ loss and similarity matrix}

\section{Evaluation of Supervised Discrete Hashing with Deep Neural Network}
\label{sec:eva_sdh-dnn}
\begin{figure*}[!t]
\centering
\subfigure[]{
       \includegraphics[scale=0.35]{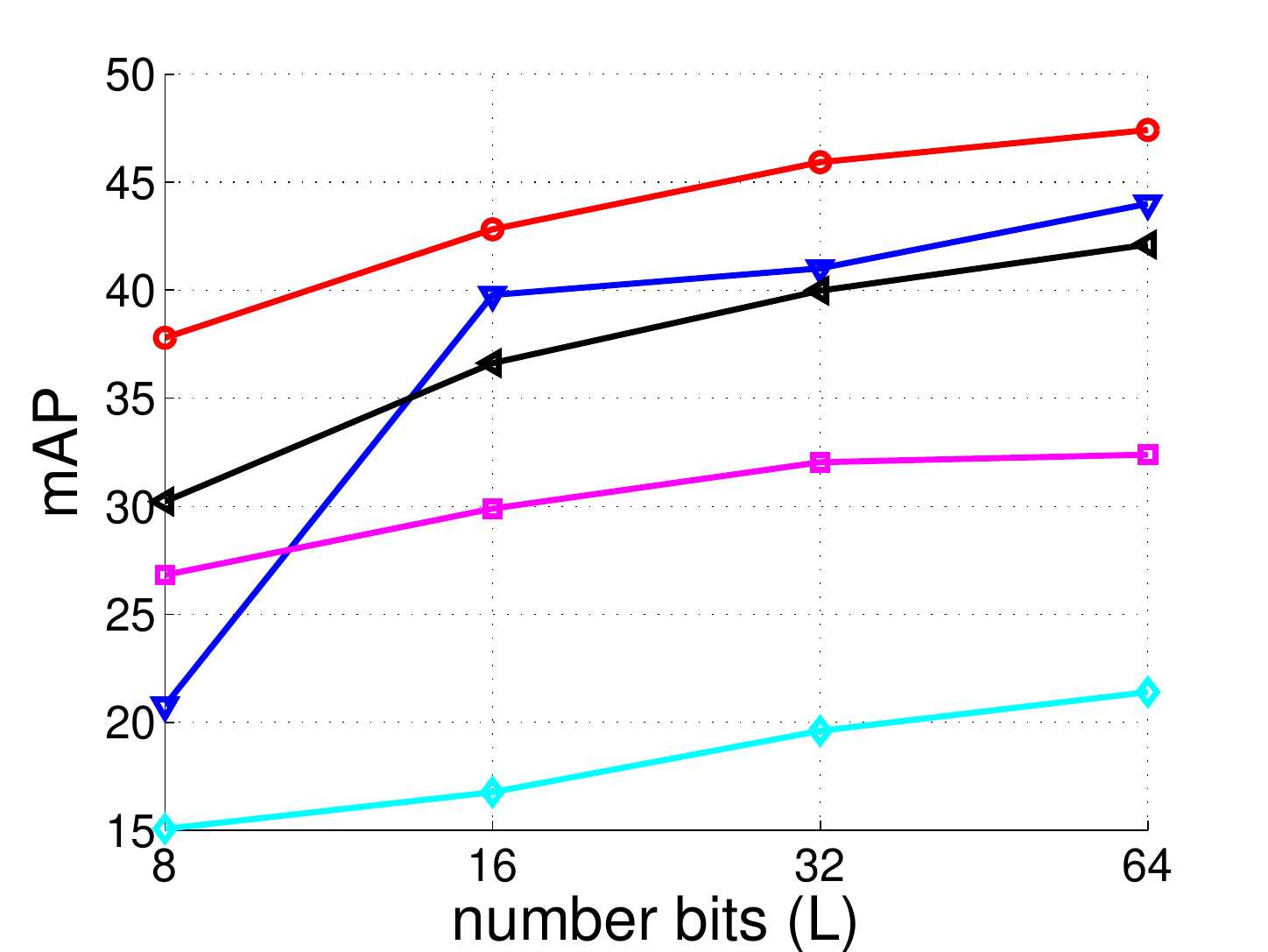}
       \label{fig:cifar_mAP_sup}
}
\subfigure[]{
       \includegraphics[scale=0.35]{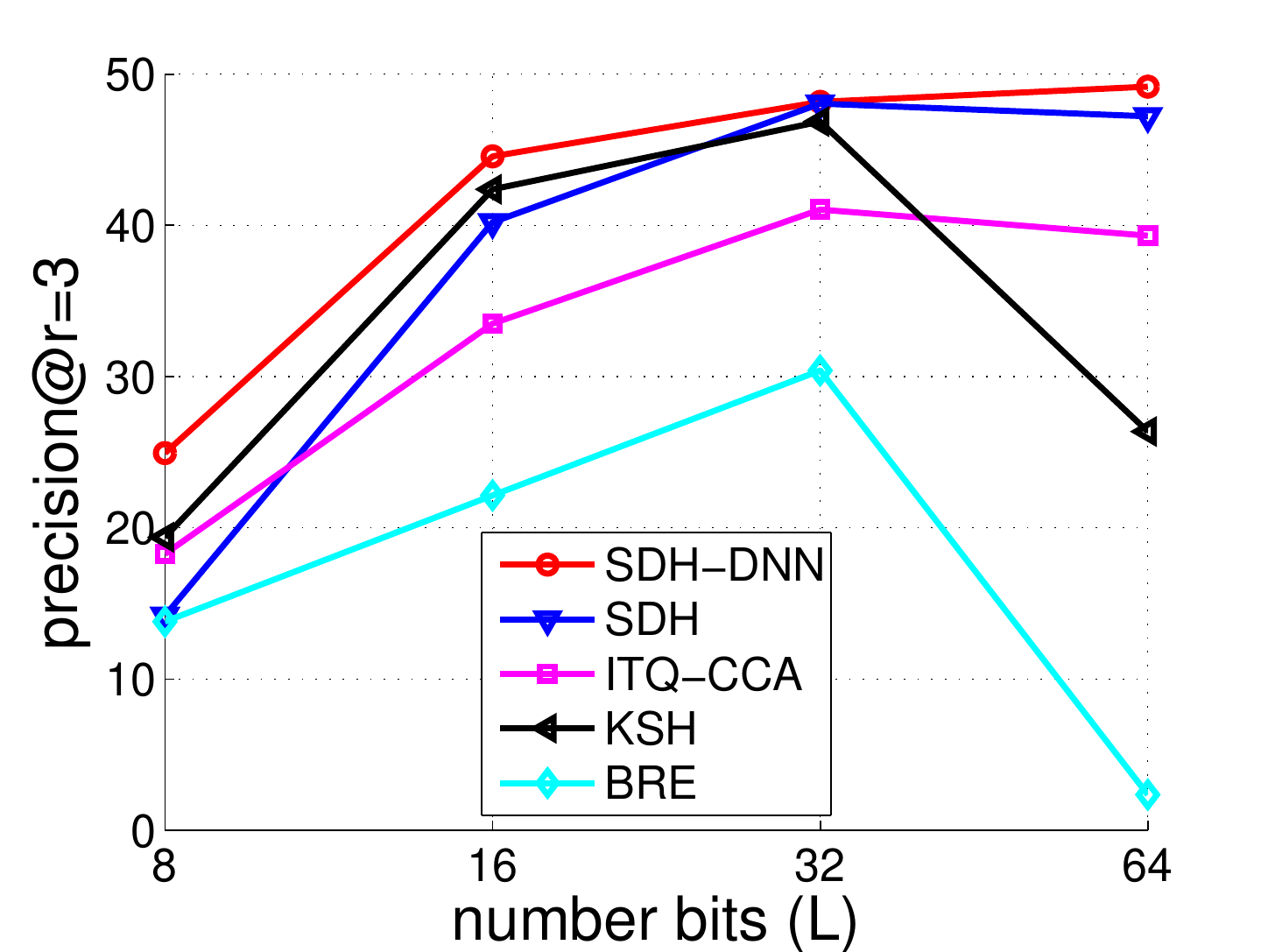} 
       \label{fig:cifar_pr3_sup}
}
\subfigure[]{
       \includegraphics[scale=0.35]{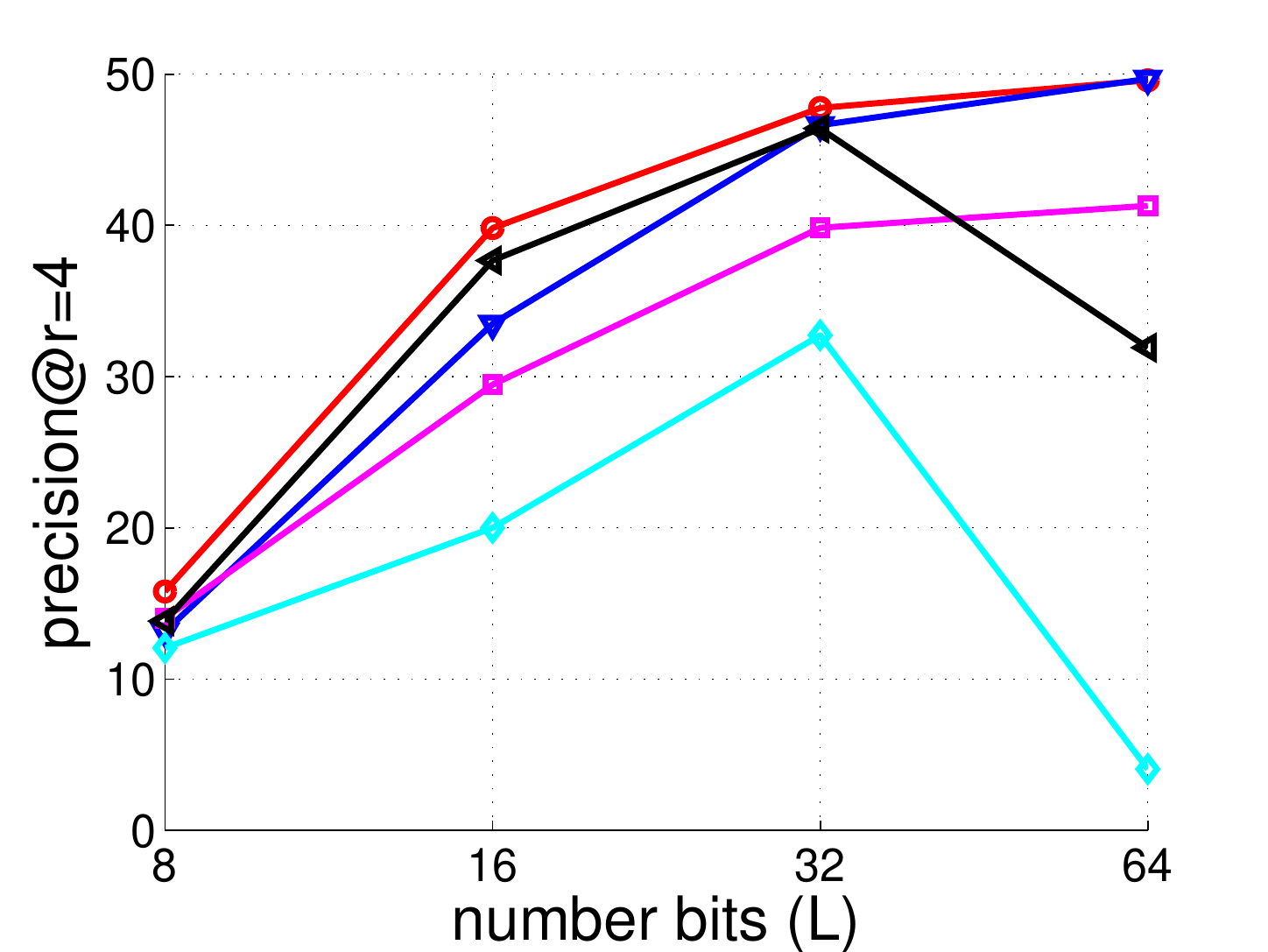} 
       \label{fig:cifar_pr4_sup}
}
\caption[]{Comparison between SDH-DNN and the state of the art on CIFAR-10 dataset.~\ref{fig:cifar_mAP_sup}: mAP. \ref{fig:cifar_pr3_sup} and \ref{fig:cifar_pr4_sup}: Precision when considering retrieved images with in Hamming distance 3 and 4, respectively.}
\label{fig:cifar10_sup}
\end{figure*}

\begin{figure*}[!t]
\centering
\subfigure[]{
       \includegraphics[scale=0.35]{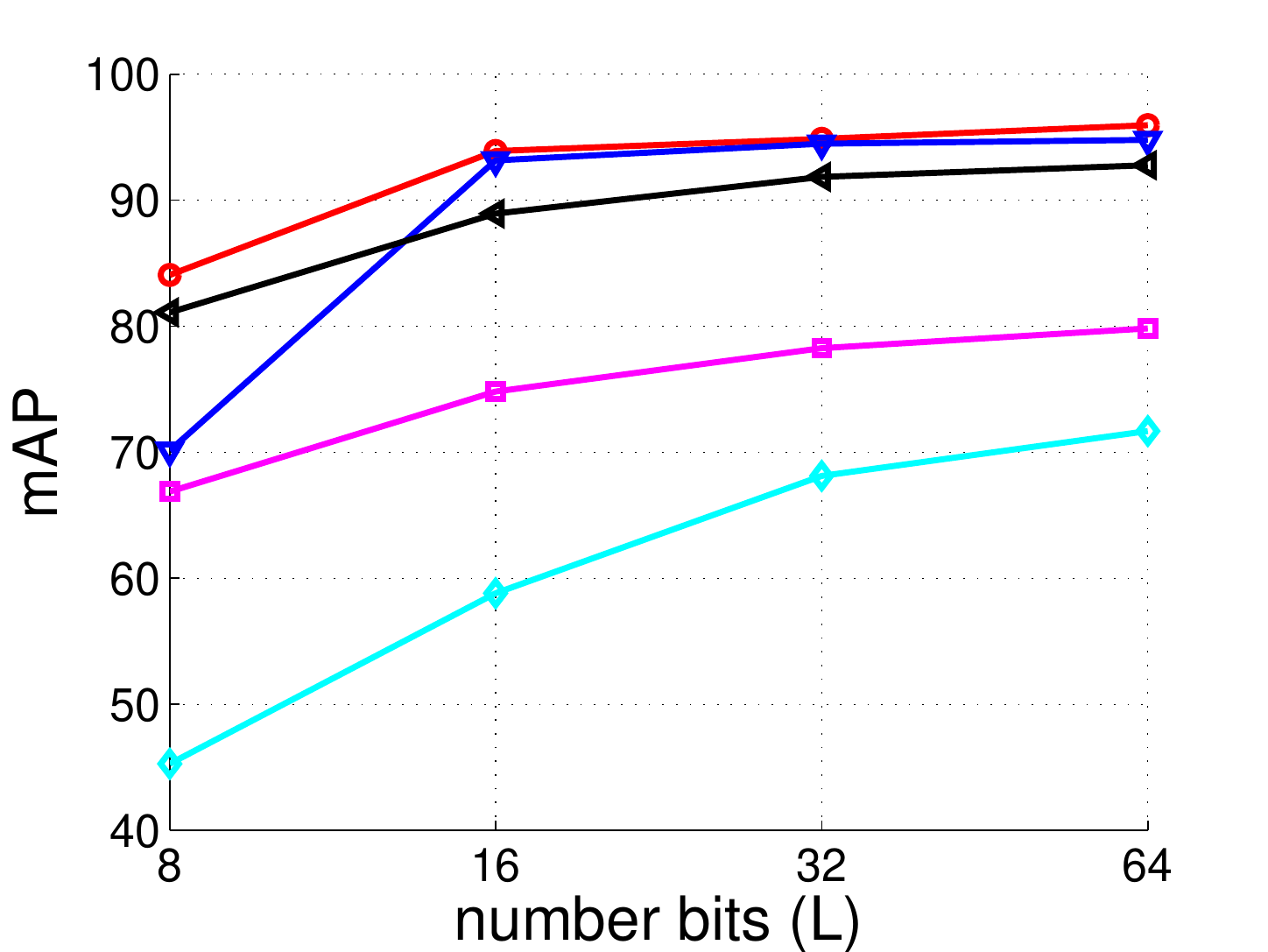}
       \label{fig:mnist_mAP_sup}
}
\subfigure[]{
       \includegraphics[scale=0.35]{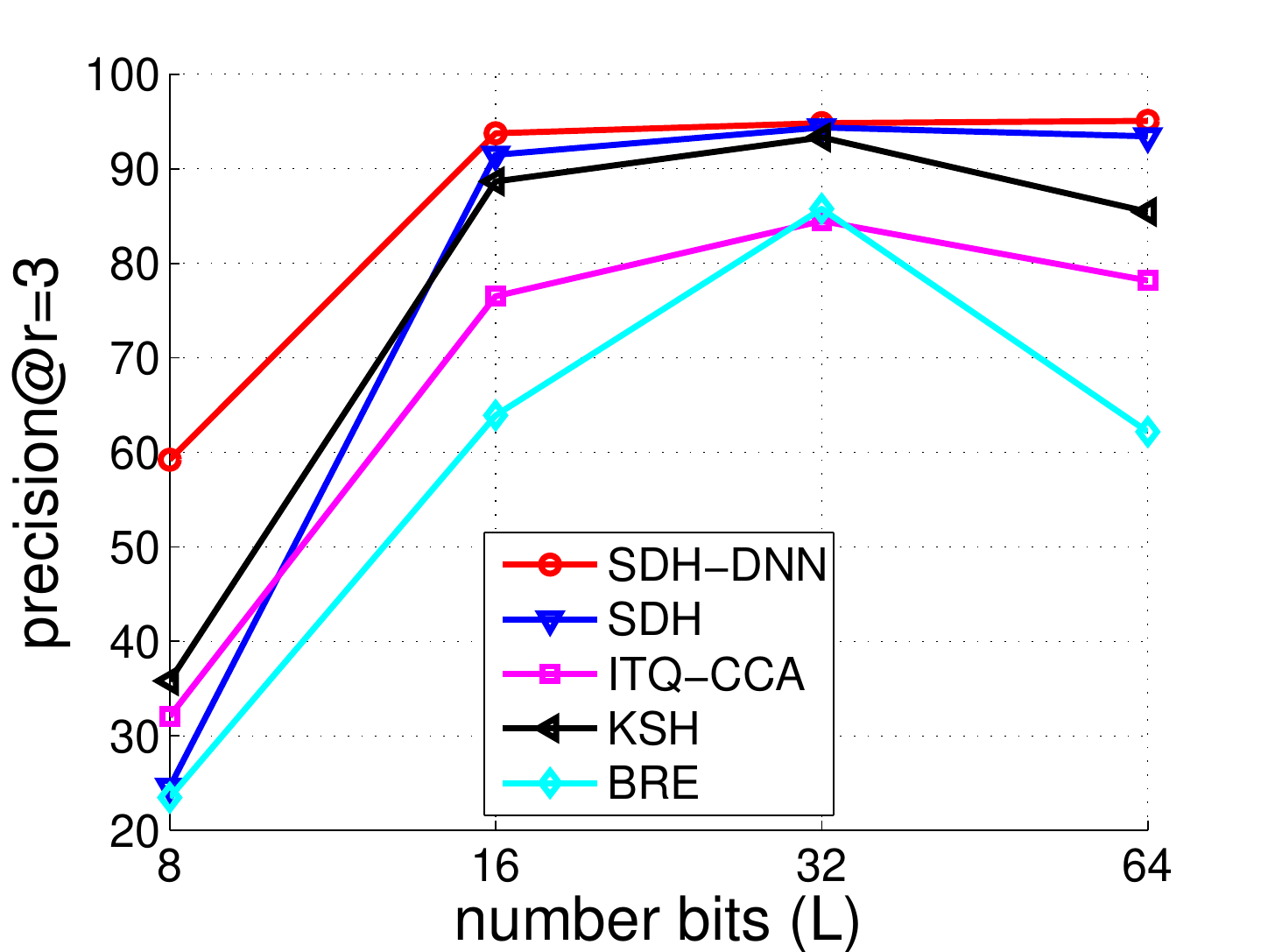} 
       \label{fig:mnist_pr3_sup}
}
\subfigure[]{
       \includegraphics[scale=0.35]{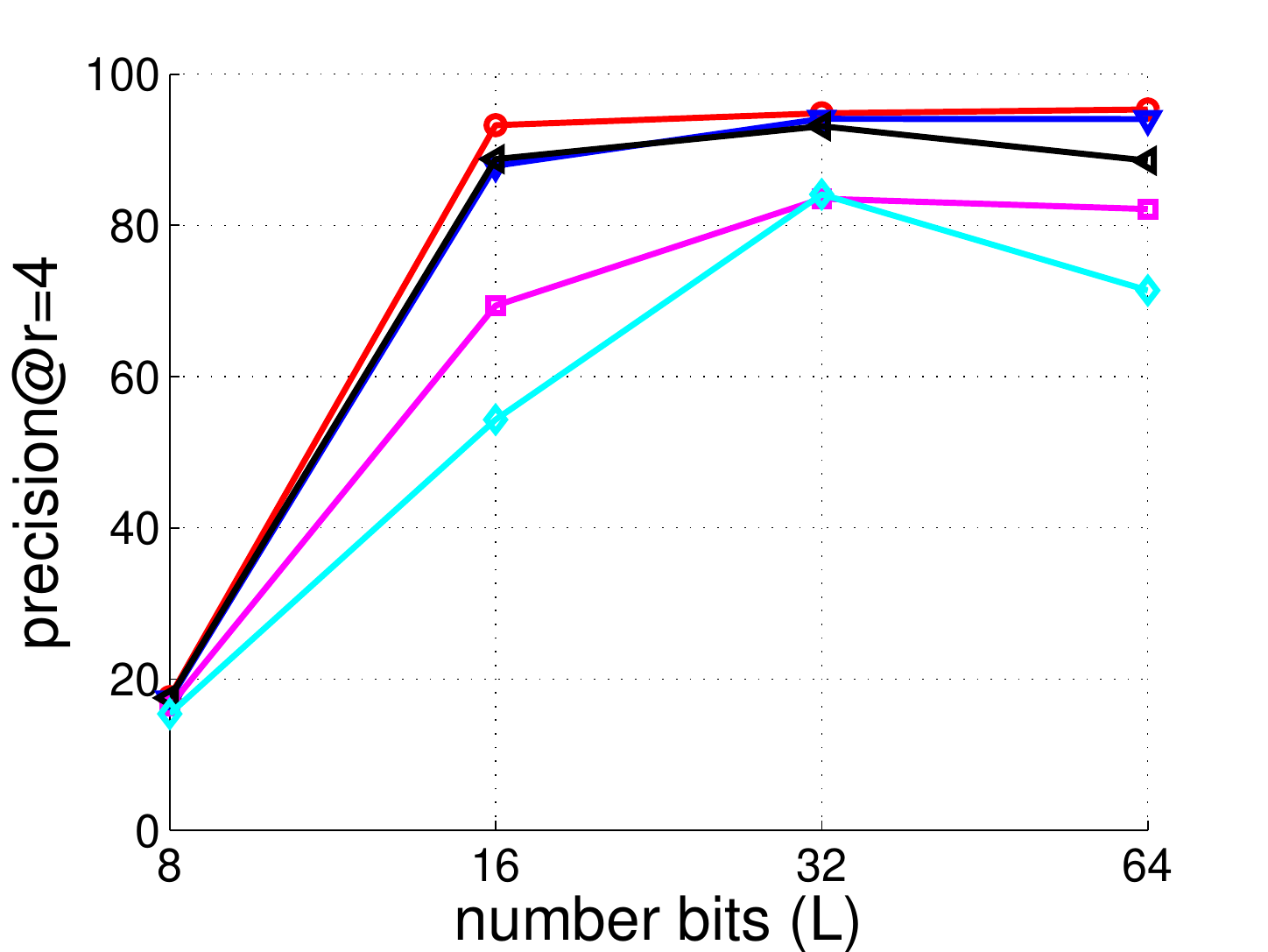} 
       \label{fig:mnist_pr4_sup}
}
\caption[]{Comparison between SDH-DNN and the state of the art on MNIST dataset.~\ref{fig:mnist_mAP_sup}: mAP. \ref{fig:mnist_pr3_sup} and \ref{fig:mnist_pr4_sup}: Precision when considering retrieved images with in Hamming distance 3 and 4, respectively.}
\label{fig:mnist_sup}
\end{figure*}
This section evaluates the proposed SDH-DNN method. The proposed SDH-DNN is compared against several state-of-the-art supervised hashing methods including Supervised Discrete Hashing (SDH)~\cite{Shen_2015_CVPR}, ITQ-CCA~\cite{DBLP:conf/cvpr/GongL11}, KSH~\cite{CVPR12:Hashing}, BRE~\cite{Kulis_learningto}. For all compared methods, we use the codes and the suggested parameters provided by the authors.
\subsection{Dataset, Implementation note and Evaluation protocol}
\paragraph{Dataset} We evaluate the proposed methods on two widely used datasets: CIFAR-10 and MNIST. The description of these dataset is provided in section~\ref{subsec:data-imp-eva}.

\paragraph{Implementation note} The network configuration is same as UDH-DNN excepting the final layer is removed. The values of parameters $\lambda_1$, $\lambda_2$, $\lambda_3$ and $\lambda_4$ are empirically set as $10^{-3}$, $5$, $1$ and $10^{-4}$, respectively. The max iteration number $max\_iter$ is set to 5. 

For ITQ\_CCA~\cite{DBLP:conf/cvpr/GongL11} and SDH~\cite{Shen_2015_CVPR}, all training samples are used for training. For SDH-DNN, KSH~\cite{CVPR12:Hashing}, BRE~\cite{Kulis_learningto} which label information is leveraged by pairwise label matrix $\S$, we randomly select $2,000$ training samples from each class and use these selected samples as new training set. The pairwise label matrix $\S$ in SDH-DNN is immediately obtained by using~(\ref{eq:S}) because the exact labels are available. 

\paragraph{Evaluation protocal} Follow standard setting for evaluating supervised hashing methods~\cite{Shen_2015_CVPR,DBLP:conf/cvpr/GongL11}, we report the retrieval results in two metrics 1) mean average precision (mAP) and 2) precision of Hamming radius $r$ (precision$@r$) which measure precision on retrieved images having Hamming distance to query $\le r$ (if no images satisfy, we report zero precision). As standardly done in the literature~\cite{Shen_2015_CVPR,DBLP:conf/cvpr/GongL11}, the ground truths are defined by the class labels from the datasets. 

\subsection{Retrieval results}

\subsubsection{Results on CIFAR-10} Figure~\ref{fig:cifar10_sup} shows comparative results on CIFAR-10 dataset. In term of mAP, we can clearly see that the proposed SDH-DNN outperforms all compared methods by a fair margin on all code lengths. 
%The improvement of SDH-DNN over the current state-of-the-art SDH~\cite{Shen_2015_CVPR} is \textbf{+8.5\%}, \textbf{+4.3\%}, \textbf{+5\%} and \textbf{+3.8\%} at 8, 16, 32 and 64 bits, respectively. The improvements of SDH-DNN over KSH~\cite{CVPR12:Hashing} which also uses label matrix are \textbf{+7.6\%}, \textbf{+5.3\%}, \textbf{+5.1\%} and \textbf{+4.1\%} at 8, 16, 32 and 64 bits, respectively.
The improvement of SDH-DNN over the current state-of-the-art supervised hashing SDH~\cite{Shen_2015_CVPR} is \textbf{+17\%}, \textbf{+3.1\%}, \textbf{+4.9\%} and \textbf{+3.4\%} at 8, 16, 32 and 64 bits, respectively. The improvements of SDH-DNN over KSH~\cite{CVPR12:Hashing} which also uses pairwise label matrix are \textbf{+7.6\%}, \textbf{+6.2\%}, \textbf{+5.9\%} and \textbf{+5.3\%} at 8, 16, 32 and 64 bits, respectively.

In term of precision of Hamming radius, the proposed SDH-DNN clearly outperforms the compared methods at low code lengths, i.e., $L = 8,16$. SDH~\cite{Shen_2015_CVPR} becomes comparable with SDH-DNN when increasing the code lengths, i.e., $L=32, 64$.

\subsubsection{Results on MNIST} Figure~\ref{fig:mnist_sup} shows comparative results on MNIST dataset. In term of mAP, the proposed SDH-DNN outperforms the current state-of-the-art SDH \textbf{+13.9\%} at $L=8$ bits. When $L$ increases, SDH-DNN and SDH~\cite{Shen_2015_CVPR} achieve similar performance. In comparison with KSH~\cite{CVPR12:Hashing}, SDH-DNN significantly outperforms KSH at all code lengths; the improvements are \textbf{+3\%}, \textbf{+4.9\%}, \textbf{+3\%} and \textbf{+3.2\%} at 8, 16, 32 and 64 bits, respectively.  

In term of precision of Hamming radius, the SDH-DNN show a clearly improvement over SDH~\cite{Shen_2015_CVPR} when $r=3$ and $L=8$. 
%At other settings, i.e. $r=3$ and $L=16, 32, 64$ or $r=4$, SDH-DNN and SDH~\cite{Shen_2015_CVPR} achieve similar performance. 
At other settings, SDH-DNN and SDH~\cite{Shen_2015_CVPR} achieve similar performance. 

\section{Conclusion}
\label{sec:conclusion}
In this paper, we propose two novel hashing methods that are UDH-DNN for unsupervised hashing and SDH-DNN for supervised hashing for learning compact binary codes. Our methods include all necessary criteria for producing good binary codes such as similarity preserving, independent and balancing. Another advantage of proposed methods are that the binary constraint on codes are directly solved during optimization without any relaxation. The experimental results on three benchmark datasets show the proposed methods compare favorably with state-of-the-art hashing methods.
{\small
\bibliographystyle{ieee}
\bibliography{hash}

\begin{thebibliography}{10}\itemsep=-1pt

\bibitem{Ng_deep}
N.~Andrew.
\newblock {Multi-Layer Neural Network}.
\newblock
  \url{http://ufldl.stanford.edu/tutorial/supervised/MultiLayerNeuralNetworks/}.

\bibitem{BA_CVPR15}
M.~A. Carreira-Perpinan and R.~Raziperchikolaei.
\newblock Hashing with binary autoencoders.
\newblock In {\em CVPR}, 2015.

\bibitem{DBLP:journals/csur/DattaJLW08}
R.~Datta, D.~Joshi, J.~Li, and J.~Z. Wang.
\newblock Image retrieval: Ideas, influences, and trends of the new age.
\newblock {\em {ACM} Comput. Surv.}, 2008.

\bibitem{Liong_2015_CVPR}
V.~Erin~Liong, J.~Lu, G.~Wang, P.~Moulin, and J.~Zhou.
\newblock Deep hashing for compact binary codes learning.
\newblock In {\em CVPR}, 2015.

\bibitem{lsh_vldb09}
A.~Gionis, P.~Indyk, and R.~Motwani.
\newblock Similarity search in high dimensions via hashing.
\newblock In {\em VLDB}, 1999.

\bibitem{DBLP:conf/cvpr/GongL11}
Y.~Gong and S.~Lazebnik.
\newblock Iterative quantization: {A} procrustean approach to learning binary
  codes.
\newblock In {\em CVPR}, 2011.

\bibitem{DBLP:journals/pami/GongLGP13}
Y.~Gong, S.~Lazebnik, A.~Gordo, and F.~Perronnin.
\newblock Iterative quantization: {A} procrustean approach to learning binary
  codes for large-scale image retrieval.
\newblock {\em PAMI}, pages 2916--2929, 2013.

\bibitem{Grauman_review}
K.~Grauman and R.~Fergus.
\newblock Learning binary hash codes for large-scale image search.
\newblock {\em Machine Learning for Computer Vision}, 2013.

\bibitem{DBLP:conf/cvpr/HeWS13}
K.~He, F.~Wen, and J.~Sun.
\newblock K-means hashing: An affinity-preserving quantization method for
  learning binary compact codes.
\newblock In {\em CVPR}, 2013.

\bibitem{CVPR12:SphericalHashing}
J.-P. Heo, Y.~Lee, J.~He, S.-F. Chang, and S.-e. Yoon.
\newblock Spherical hashing.
\newblock In {\em CVPR}, 2012.

\bibitem{herve_pami2011}
H.~J{\'e}gou, M.~Douze, and C.~Schmid.
\newblock Product quantization for nearest neighbor search.
\newblock {\em PAMI}, pages 117--128, 2011.

\bibitem{conf/nips/KongL12}
W.~Kong and W.-J. Li.
\newblock Isotropic hashing.
\newblock In {\em NIPS}, 2012.

\bibitem{Krizhevsky09}
A.~Krizhevsky.
\newblock Learning multiple layers of features from tiny images.
\newblock Technical report, University of Toronto, 2009.

\bibitem{Kulis_learningto}
B.~Kulis and T.~Darrell.
\newblock Learning to hash with binary reconstructive embeddings.
\newblock In {\em NIPS}, 2009.

\bibitem{KLSH_iccv09}
B.~Kulis and K.~Grauman.
\newblock Kernelized locality-sensitive hashing for scalable image search.
\newblock In {\em ICCV}, 2009.

\bibitem{DBLP:journals/pami/KulisJG09}
B.~Kulis, P.~Jain, and K.~Grauman.
\newblock Fast similarity search for learned metrics.
\newblock {\em PAMI}, pages 2143--2157, 2009.

\bibitem{mnistlecun}
Y.~Lecun and C.~Cortes.
\newblock {The MNIST database of handwritten digits}.
\newblock \url{http://yann.lecun.com/exdb/mnist/}.

\bibitem{CVPR2014Lin}
G.~Lin, C.~Shen, Q.~Shi, A.~{van den Hengel}, and D.~Suter.
\newblock Fast supervised hashing with decision trees for high-dimensional
  data.
\newblock In {\em CVPR}, 2014.

\bibitem{Liu89onthe}
D.~C. Liu and J.~Nocedal.
\newblock On the limited memory bfgs method for large scale optimization.
\newblock {\em Mathematical Programming}, 45:503--528, 1989.

\bibitem{CVPR12:Hashing}
W.~Liu, J.~Wang, R.~Ji, Y.-G. Jiang, and S.-F. Chang.
\newblock Supervised hashing with kernels.
\newblock In {\em CVPR}, 2012.

\bibitem{DBLP:conf/icml/LiuWKC11}
W.~Liu, J.~Wang, S.~Kumar, and S.~Chang.
\newblock Hashing with graphs.
\newblock In {\em ICML}, 2011.

\bibitem{malick:hal-00389552}
J.~Malick, J.~Povh, F.~Rendl, and A.~Wiegele.
\newblock {Regularization Methods for Semidefinite Programming}.
\newblock {\em {SIAM Journal on Optimization}}, pages 336--356, 2009.

\bibitem{minhdo_hash2014}
V.~A. Nguyen, J.~Lu, and M.~N. Do.
\newblock Supervised discriminative hashing for compact binary codes.
\newblock In {\em ACM MM}, 2014.

\bibitem{lbfgs2}
J.~Nocedal.
\newblock {Updating Quasi-Newton Matrices with Limited Storage}.
\newblock {\em Mathematics of Computation}, pages 773--782, 1980.

\bibitem{DBLP:conf/icml/NorouziF11}
M.~Norouzi and D.~J. Fleet.
\newblock Minimal loss hashing for compact binary codes.
\newblock In {\em ICML}, 2011.

\bibitem{DBLP:conf/nips/0002FS12}
M.~Norouzi, D.~J. Fleet, and R.~Salakhutdinov.
\newblock Hamming distance metric learning.
\newblock In {\em NIPS}, 2012.

\bibitem{gist}
A.~Oliva and A.~Torralba.
\newblock Modeling the shape of the scene: A holistic representation of the
  spatial envelope.
\newblock {\em IJCV}, pages 145--175, 2001.

\bibitem{KLSH_nips09}
M.~Raginsky and S.~Lazebnik.
\newblock Locality-sensitive binary codes from shift-invariant kernels,”
  advances in neural information processing systems, 2009.

\bibitem{DBLP:SeH}
R.~Salakhutdinov and G.~E. Hinton.
\newblock Semantic hashing.
\newblock {\em Int. J. Approx. Reasoning}, pages 969--978, 2009.

\bibitem{Shen_2015_CVPR}
F.~Shen, C.~Shen, W.~Liu, and H.~Tao~Shen.
\newblock Supervised discrete hashing.
\newblock In {\em CVPR}, June 2015.

\bibitem{DBLP:journals/pami/StrechaBBF12}
C.~Strecha, A.~M. Bronstein, M.~M. Bronstein, and P.~Fua.
\newblock Ldahash: Improved matching with smaller descriptors.
\newblock {\em PAMI}, pages 66--78, 2012.

\bibitem{DBLP:journals/pami/WangKC12}
J.~Wang, S.~Kumar, and S.~Chang.
\newblock Semi-supervised hashing for large-scale search.
\newblock {\em PAMI}, pages 2393--2406, 2012.

\bibitem{DBLP:journals/corr/WangSSJ14}
J.~Wang, H.~T. Shen, J.~Song, and J.~Ji.
\newblock Hashing for similarity search: {A} survey.
\newblock {\em CoRR}, 2014.

\bibitem{DBLP:conf/nips/WeissTF08}
Y.~Weiss, A.~Torralba, and R.~Fergus.
\newblock Spectral hashing.
\newblock In {\em NIPS}, 2008.

\end{thebibliography}
}

\end{document}